\documentclass[12pt,twoside]{report}

\usepackage{hyperref}
\hypersetup{
    hyperindex, linktoc=page, breaklinks,
    pdfauthor={Carlos Garcia-Saura},pdftitle={Self-calibration of a differential wheeled robot using only a gyroscope and a distance sensor}
}

\usepackage[utf8]{inputenc}
\usepackage{amsmath}
\usepackage{amsfonts}
\usepackage{amssymb}

\usepackage{setspace}

\usepackage{enumitem}

\usepackage[normalem]{ulem}

\usepackage[table,xcdraw]{xcolor}

\newcommand{\reporttitle}{Self-calibration of a differential wheeled robot using only a gyroscope and a distance sensor}
\newcommand{\reportauthor}{Carlos Garcia-Saura}
\newcommand{\supervisor}{Prof. Andrew J. Davison}
\newcommand{\degreetype}{Computing Science (Specialism in A.I.)}

\newcommand{\keywords}{Mobile robotics, differential wheeled robot, motion control, self-calibration, continuous-rotation servomotor, PID auto-tuning, multimodal sensing, MEMS gyroscope, IR distance sensor, open-source}

\usepackage[a4paper,left=2.8cm,right=2.8cm,vmargin=2.0cm,includeheadfoot]{geometry}
\usepackage{textpos}
\usepackage{tabularx,longtable,multirow,subfigure,caption}%hangcaption
\usepackage{fncylab} %formatting of labels
\usepackage{fancyhdr} % page layout
\usepackage{url} % URLs
\usepackage[english]{babel}
\usepackage{amsmath}
\usepackage{graphicx}
\usepackage{dsfont}
\usepackage{array}
\usepackage{latexsym}
\def\@makechapterhead#1{%
  \vspace*{10\p@}%
  {\parindent \z@ \raggedright \sffamily
    \interlinepenalty\@M
    \Huge\bfseries \thechapter \space\space #1\par\nobreak
    \vskip 30\p@
  }}

\def\@makeschapterhead#1{%
  \vspace*{10\p@}%
  {\parindent \z@ \raggedright
    \sffamily
    \interlinepenalty\@M
    \Huge \bfseries  #1\par\nobreak
    \vskip 30\p@
  }}

\allowdisplaybreaks

\date{September 2015}

\begin{document}

% load title page
% Last modification: 2015-08-17 (Marc Deisenroth)
\begin{titlepage}

\newcommand{\HRule}{\rule{\linewidth}{0.5mm}} % Defines a new command for the horizontal lines, change thickness here

%----------------------------------------------------------------------------------------
%	LOGO SECTION
%----------------------------------------------------------------------------------------

\includegraphics[width = 4cm]{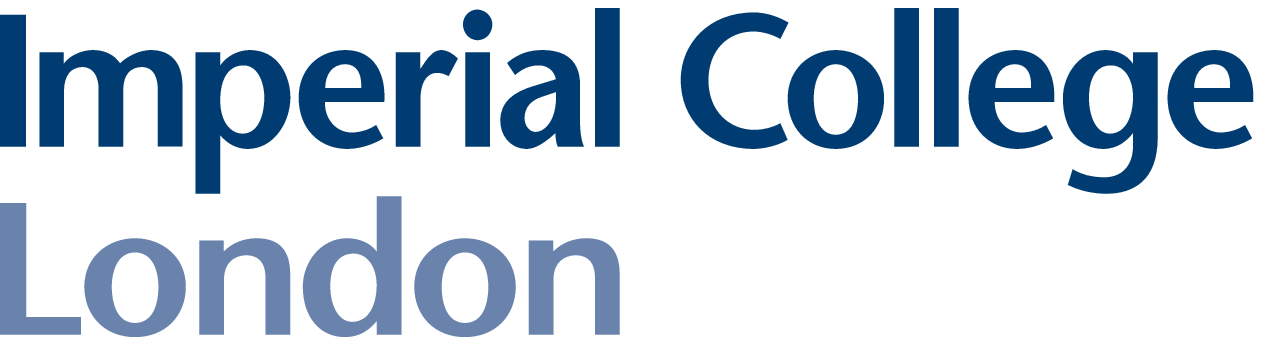}\\[0.5cm] 

\center % Center remainder of the page

%----------------------------------------------------------------------------------------
%	HEADING SECTIONS
%----------------------------------------------------------------------------------------

\textsc{\Large Imperial College London}\\[0.5cm] 
\textsc{\large Department of Computing}\\[0.5cm] 

%----------------------------------------------------------------------------------------
%	TITLE SECTION
%----------------------------------------------------------------------------------------

\HRule \\[0.4cm]
%{ \huge \bfseries \reporttitle}\\ % Title of your document
\begin{spacing}{1.5}
{ \fontsize{0.65cm}{1em} \bfseries \reporttitle}\\ % Title of your document
\end{spacing}
\HRule \\[1.5cm]
 
%----------------------------------------------------------------------------------------
%	AUTHOR SECTION
%----------------------------------------------------------------------------------------

\begin{minipage}{0.4\textwidth}
\begin{flushleft} \large
\emph{Author:}\\
\reportauthor % Your name
\end{flushleft}
\end{minipage}
~
\begin{minipage}{0.4\textwidth}
\begin{flushright} \large
\emph{Supervisor:} \\
\supervisor % Supervisor's Name
\end{flushright}
\end{minipage}\\[4cm]

%----------------------------------------------------------------------------------------
%	FOOTER & DATE SECTION
%----------------------------------------------------------------------------------------
\vfill % Fill the rest of the page with whitespace
Submitted in partial fulfillment of the requirements for the MSc degree in
\degreetype~of Imperial College London\\[0.5cm]

\makeatletter
\@date 
\makeatother

\end{titlepage}

\clearpage{\pagestyle{empty}\cleardoublepage}

% page numbering etc.
\pagenumbering{roman}
%\clearpage{\pagestyle{empty}\cleardoublepage}
\setcounter{page}{1}
\pagestyle{fancy}

%%%%%%%%%%%%%%%%%%%%%%%%%%%%%%%%%%%%
\begin{abstract}

Research in mobile robotics often demands platforms that have an adequate balance between cost and reliability. In the case of terrestrial robots, 
one of the available options is the GNBot, an open-hardware project intended for the evaluation of swarm search strategies.
The lack of basic odometry sensors such as wheel encoders had so far difficulted the implementation of an accurate high-level controller in this platform.
Thus, the aim of this thesis is to improve motion control in the GNBot by incorporating a gyroscope whilst maintaining the requisite of no wheel encoders.
Among the problems that have been tackled are: accurate in-place rotations, minimal drift during linear motions, and arc-performing functionality.
Additionally, the resulting controller is calibrated autonomously by using both the gyroscope module and the infrared rangefinder on board each robot, greatly simplifying the calibration of large swarms.
The report first explains the design decisions that were made in order to implement the self-calibration routine, and then evaluates the performance of the new motion controller by means of off-line video tracking. The motion accuracy of the new controller is also compared with the previously existing solution in an odor search experiment.

\end{abstract}

\clearpage{\pagestyle{empty}\cleardoublepage}
%%%%%%%%%%%%%%%%%%%%%%%%%%%%%%%%%%%%
\section*{Acknowledgements}
\emph{In first place I must thank professor Andrew J. Davison at Imperial College London for his patient guidance and for pointing this project in the right direction.
I found the Robotics course to be very inspiring so I want to thank as well the rest of coordinators and course assistants.}

\emph{I am also thankful to professors Pablo Varona and Francisco de Borja Rodr\'iguez at Universidad Aut\'onoma de Madrid, for allowing me to use their infrastructure to perform the experiments. My friends Alejandro, Irene and Aar\'on also deserve a mention for making the research time in the lab (and outside the lab!) way more enjoyable.}

\emph{From the great people I have met this year at Imperial College London, I want to mention in special Hadeel, Bob, Matthew, Stefan, Marta, \sout{Dony}Doniyor, Pawel, Tony, Siwook, Yui, and Win for always being there, for the good and the bad, no matter how complicated the assignment or big the cake! :-)
I must mention as well the IC Robotics Society and IC Advanced Hackspace, in special Larissa and Audrey for helping a lot with the 3D printers and laser cutters.
Also, my year in London wouldn't have been the same without yOPERO, Erin, Charlaine, Christian and Cliona.}

\emph{I also want to thank the whole open-source community (specially GNU/Linux, Python, Numpy, Matplotlib, OpenCV, Arduino, RepRap, etc) for making it possible to perform this project without requiring any proprietary software or machinery.}

\emph{Finally, big thanks go to my parents for their continuous loving support, and also thanks to Rodrigo, FJ and Sandra for tolerating my mood swings while working in this project during our family holidays.}

\emph{!`Muchas gracias a todos!}

\emph{-- Carlos}

\clearpage{\pagestyle{empty}\cleardoublepage}

%%%%%%%%%%%%%%%%%%%%%%%%%%%%%%%%%%%%
%--- table of contents
\fancyhead[LE,RO]{\slshape}
\fancyhead[RE,LO]{\sffamily {Table of Contents}}

\begin{spacing}{0.1}
\tableofcontents 
\end{spacing}

\clearpage{\pagestyle{empty}\cleardoublepage}

\fancyhead[RE,LO]{\sffamily {List of Figures}}

\begin{spacing}{0.1}
\listoffigures
\end{spacing}

\clearpage{\pagestyle{empty}\cleardoublepage}
\pagenumbering{arabic}
\setcounter{page}{1}

\fancyhead[LE,RO]{\slshape}
%\fancyhead[LE,RO]{\slshape \rightmark}
\fancyhead[LO,RE]{\slshape \leftmark}

%%%%%%%%%%%%%%%%%%%%%%%%%%%%%%%%%%%%
\chapter{Introduction}

\begin{figure}[hbtp]
\centerline{
\includegraphics[width=0.8\linewidth]{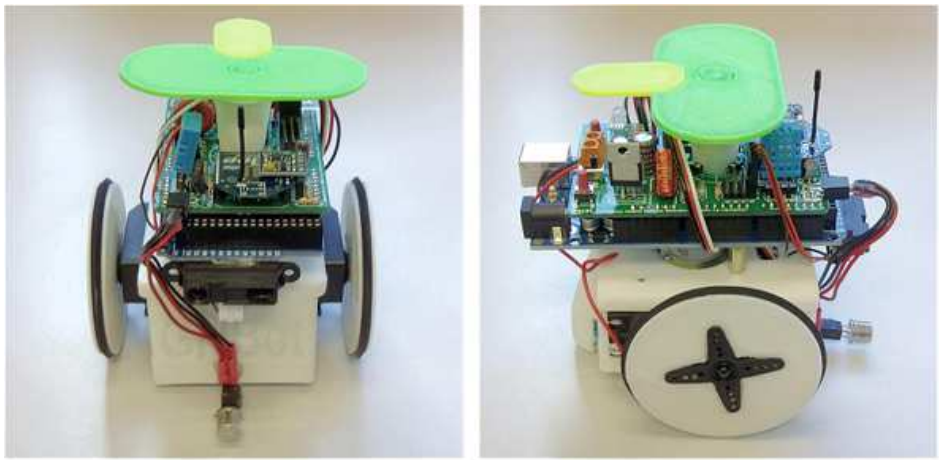}}
\caption[One of the robots from the GNBot swarm used for the project]{\emph{One of the robots from the GNBot swarm used for the project.}
The electronics are based on the Arduino MEGA board. A custom shield contains all of the multimodal sensors, which provide the robot with gas sensing capabilities (TGS-2900), as well as distance sensing (GP2Y0A21YK0F), temperature \& humidity (DHT11), and light intensity sensing.
\textbf{This project additionally incorporated a gyroscope module (MPU6050)}.
The main actuators are two continuous-rotation servomotors, and each robot also has a multicolor RGB LED and a piezoelectric speaker.
Finally, the wireless communication layer is based on ZigBee, and the green/yellow top markers allow for external video tracking of the swarm.
From \cite{GarciaSauraLM14,GarciaSaura14}.
}
\label{fig:GNBotViews}
\end{figure}

The starting point of this project is the \emph{GNBot} swarm robot platform that was presented in \emph{``Design principles for cooperative robots with uncertainty-aware and resource-wise adaptive behavior''}\cite{GarciaSauraLM14}, and \emph{``Cooperative strategies for the detection and localization of odorants with robots and artificial noses''}\cite{GarciaSaura14}.
Figure \ref{fig:GNBotViews} describes the most relevant characteristics of these robots.

The GNBot is a \emph{differential wheeled robot}, which means that it has two active wheels sharing the same axis, and each of them is actuated independently. There is also a passive, free turning ball caster that acts as a third stand. If the desired trajectory is a straight path, the velocities are set to equal magnitude; if a rotation is required instead, then different velocities can be applied to each motor.

Using this setup it is possible to estimate the motion of the robot in the XY plane given its starting position ($P_{xy\alpha}$), the rotational velocity of each motor ($\omega_L$, $\omega_R$), the diameter of the wheels ($D_L$, $D_R$) and their separation ($d$). This estimation process is known as \emph{dead reckoning}\footnote{\url{https://en.wikipedia.org/wiki/Dead_reckoning}}.

\begin{figure}[hbtp]
\centerline{
\includegraphics[width=0.85\linewidth]{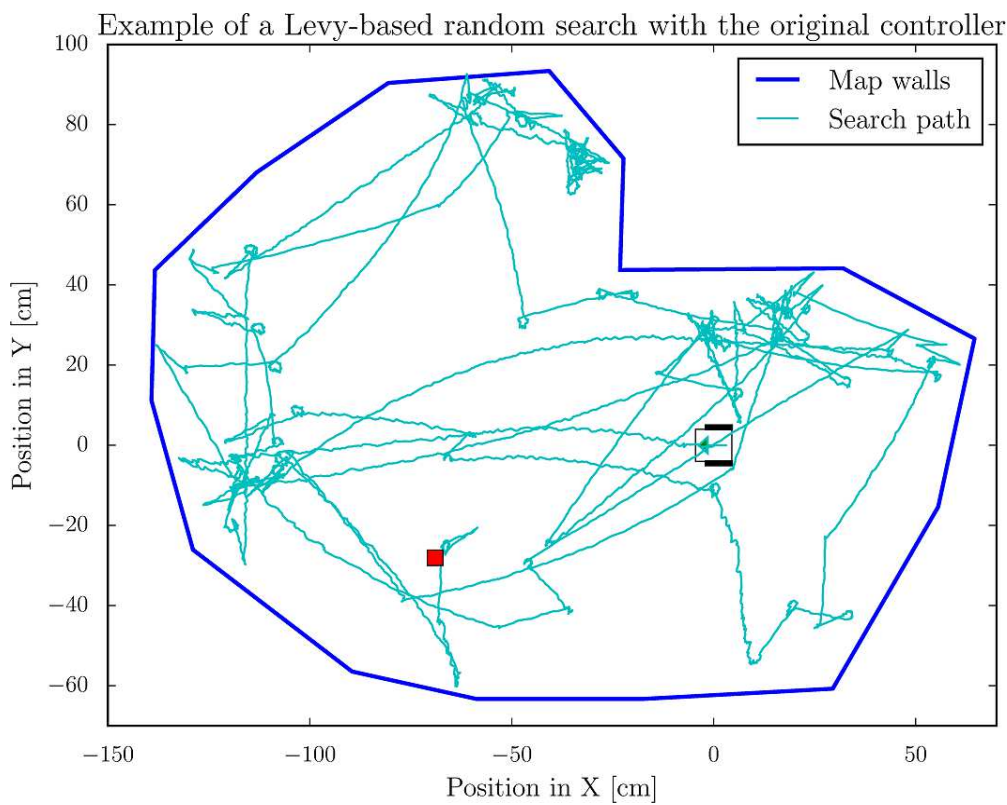}}
\caption[Example of an odor search strategy based on L\'{e}vy walks, as performed by the original robot controller]{\emph{Example of an odor search strategy based on L\'{e}vy walks, as performed by the original robot controller.}
The search is defined at a high level as random in-place rotations combined with linear trajectories whose lengths vary according to a heavy-tailed probabilistic distribution. \textbf{It can be appreciated that the segments that were supposed to be linear, are actually performed as arcs (there is drift in yaw)}.
In this experiment, the target odor source (red marker) took 12 minutes to be found.
The ground truth trajectory was recorded with a webcam, post perspective correction and color marker tracking with the OpenCV software library.}
\label{fig:levyDriftSingle}
\end{figure}

The accuracy of dead reckoning relies on how dependable the measurements used in the calculations are. These include wheel separation and their diameter, but most importantly the actual rotational velocities of each motor.
For instance, small deviations from the theoretical velocities can turn a desirably linear trajectory into an arc. An example of this issue is displayed in Figure \ref{fig:levyDriftSingle}.

\pagebreak

This thesis has tackled the improvement of those low-level control routines in order to minimize motion drift, as well as the implementation of a distance-based abstraction layer that facilitates the use of the GNBot platform in practical applications.
It also explores the automatic calibration of the robots using only on-board sensors (a gyroscope and a distance sensor).

The following two sections are an overview of the project's approach towards the improvement the motion controller; Chapter \ref{sec:calib_chap} provides an in-depth explanation of the self-calibration algorithm; Next, Chapter \ref{sec:eval_chap} evaluates the resulting controller; Finally, Chapter \ref{sec:concl_chap} analyses the outcome of the project and suggests future research paths.

\section{Background and motivation} \label{sec:bg_section}

Undesired drift in the estimation of position, very characteristic when using dead reckoning in mobile robotics, can be reduced with the incorporation of basic \emph{odometry}\footnote{\emph{Odometry:} Use of sensory data to improve motion estimates} such as \emph{rotational encoders}\footnote{\emph{Rotational encoder:} A sensor that can measure wheel rotations accurately and in real time}.
Wheel encoders are a very common choice given their simplicity and reduced cost, but they do have some drawbacks:
\begin{itemize}[noitemsep,topsep=0pt,parsep=5pt,partopsep=0pt]
\item Incorporating wheel encoders into a previously existing design is often nontrivial, as the process involves hardware modifications  near critical moving parts.
\item Electrical, optical or magnetic wheel encoders can be susceptible to dirt; placing them near the wheels may make periodic maintenance necessary.
\item Most importantly, wheel encoders cannot easily detect whether there is any slipping between the wheel and the ground surface.
\end{itemize}

For these reasons the incorporation of wheel encoders in the GNBot has been dismissed.
Instead, the first trials towards the reduction of yaw drift included the use of an electronic compass. The goal was to obtain a global, unbiased measure of heading (yaw) by measuring the earth's magnetic field. Unfortunately the compass was deemed unusable for indoor environments during the first trials of the GNBot\cite{GarciaSaura14}, since the magnetometer measurements are substantially distorted by nearby metals (i.e. buried cables) commonly present indoors.

\begin{figure}[hbtp]
\centerline{
\includegraphics[width=0.62\linewidth]{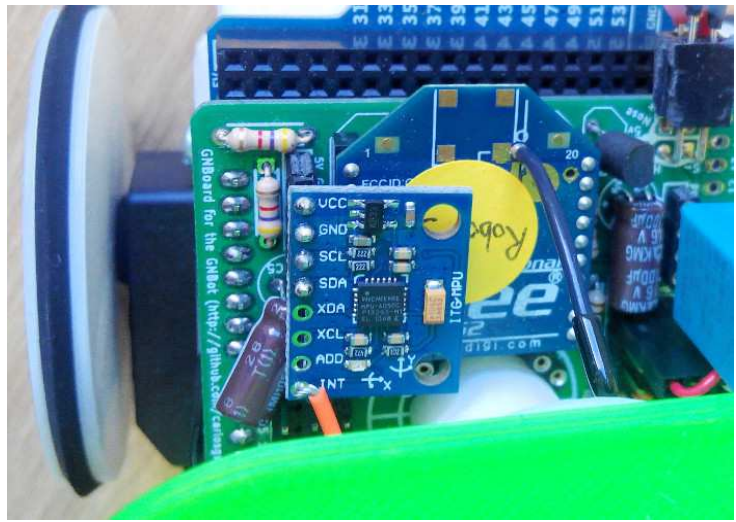}}
\caption[Detail of the gyroscope module installed in the GNBot]{\emph{Detail of the gyroscope module installed in the GNBot.} The selected board is based on the MPU6050 module, which is Arduino-compatible. The module fits on the $I^2C$ socket of the GNBot without any additional modification.
An orange wire was soldered to the interrupt pin, but it was not needed in the end, as polling was used to read values from the module.
}
\label{fig:GNBotGyro}
\end{figure}

Another option for odometry is the gyroscope, which is the approach studied in this project (see Fig. \ref{fig:GNBotGyro}). Electronic gyroscopes are a form of inertial sensor that can measure relative rotations in any orientation, without requiring interactions with external moving parts. As opposed to wheel encoders, the physical incorporation of a gyroscope into the GNBot would be quite straightforward, and would also have the advantage of being capable of easily detecting wheel slipping.

The use of inertial sensors for position estimation has been a matter of research for many decades. The last few years in particular have seen the development of smartphones, tablets, intelligent cameras and other devices in need of inertial sensors, which has greatly boosted the industry. The technology has been able to achieve high miniaturization, high accuracy and low power consumption, whilst substantially reducing costs, yielding the excellent inertial sensors that are now widely available in the market.

In robotics, Inertial Measurement Units (IMUs) are quite often employed for navigation in different environments.
For instance, the combination of IMUs with GPS allows for a remarkably accurate 3D localization in aerial robotics\cite{Barczyk10}. These techniques have also been demonstrated in ground robots without GPS\cite{PeiChunLin06,SeungbeomWoo11}, as well as in underwater robots without encoders\cite{Stilwell01,Kinsey02,Panish11}.
More recently, 3D position tracking with inertial sensors has also been employed to correct the effect of the rolling shutter of RGB\cite{HyoungKiLee12} and RGB-D\cite{Ovren13} camera sensors.

Many authors use \emph{Kalman filters}\footnote{\url{https://en.wikipedia.org/wiki/Kalman_filter}} in their motion controllers. Kalman filtering is a probabilistic method for sensor fusion, which means it can be used to efficiently combine measurements from the inertial sensor with other forms of odometry such as wheel encoders, a compass, or even barometers; this way it is possible to achieve very robust motion controllers with minimal drift\cite{PeiChunLin06,HyoungKiLee12,
HakyoungChung01,panich11}.
In fact, most of the commercially-available IMUs, such as the one selected for this project, already contain internal Kalman filters that can integrate measurements from a gyroscope, an accelerometer or a compass.

For the final evaluation of the positioning accuracy of the controller, the project took advice from work by J. Borenstein et al.\cite{HakyoungChung01,Borenstein98}, where the robots were programmed to describe square trajectories that were first logged and then compared against ground truth.

%Self calibration (SLAM):
%\cite{DeCecco02} % inertial + odometry
%\cite{roy99}
%\cite{martinelli07}

%kalman
%\cite{HakyoungChung01,PeiChunLin06,
%HyoungKiLee12,martinelli07,panich11}

%Gyro. Square for evaluation:
%\cite{Borenstein98,HakyoungChung01}

%3D navigation (including accelerometers):
%Inertial navigation systems:
%Hexapod:
%\cite{PeiChunLin06}
%UAV aided by GPS:
%\cite{Barczyk10}
%Linear velocity calibration using Fuzzy %inference:
%\cite{SeungbeomWoo11}

%Monocular SLAM:
%\cite{HyoungKiLee12}

%RGB-D rolling shutter:
%\cite{Ovren13}

%Gyro. Underwater:
%\cite{Stilwell01,Kinsey02,Panish11}

%Gyro. Kalman filter
%\cite{panich11}

\section{Contribution overview}

At the start of this project, the motion control of the GNBot robots was speed-based and open-loop (with no wheel encoders). This resulted in very inaccurate motions that drifted rapidly from the desired path. For example, when a robot was commanded to describe a straight trajectory, an arc path would be observed instead. In-place rotations were also quite inaccurate.

Thus the goal of this project has been to improve the motion accuracy of these robots and to provide a high-level distance-based controller.
In particular the following matters have been tackled:

\begin{itemize}
\item Incorporation of an inertial sensor (MPU6050) into the GNBot design, as well as the implementation of low-level functionality that reads \emph{yaw, pitch and roll} from the gyroscope module with minimal drift.
\item Analysis of the servomotor response curves. Since there are no wheel encoders, the rotations were tracked using the gyroscope, by independently rotating each wheel over the entire range of velocities.
\item Implementation of a PID controller for accurate yaw regulation; Additionally the constants are auto-calibrated with a process based on the Ziegler-Nichols method.
\item Analysis of the infrared sensor response curve, and linearisation using exponential curve fitting.
\item Analysis of the nonlinear effect of the PID yaw controller over the robot's linear velocity for the whole range.
\item The results of these analysis have been integrated into a high-level distance-based motion controller that supports linear motions and arcs. Additionally, the calibration of each robot is done autonomously by using only on-board sensors (the gyroscope and the IR rangefinder) and requiring minimal user interaction.
\item Implementation of an off-line video tracking method capable of reliably recording ground truth robot paths. The process involved perspective \& distance correction using the OpenCV library.
\item Evaluation of the new controller by making the robots perform lines, circles and squares of known dimensions; Also the system was compared against the previously existing solution in an odor search task based on a wall-bounce strategy.
\end{itemize}

The new motion controller has a linear drift in the order of $6cm/m$ in the low velocity settings ($2$ to $6cm/s$) and around $9cm/m$ at higher velocities ($6$ to $10cm/s$). The rotational drift is in the order of $2deg/min$ when the robot is moving.

In summary, this project has provided the high-level functionality needed in order to achieve accurate control over the GNBot robots. The self-calibrating nature of the approach also facilitates its use in large robot swarms.

The outcome of the project is open-source (Attribution-ShareAlike 4.0 International\footnote{\url{http://creativecommons.org/licenses/by-sa/4.0/}}) and can be accessed in the following GitHub respository:

\begin{center}
\large\url{https://github.com/carlosgs/GNBot}
\end{center}

%%%%%%%%%%%%%%%%%%%%%%%%%%%%%%%%%%%%
\chapter{The self-calibration algorithm} \label{sec:calib_chap}

\begin{figure}[hbtp]
\centerline{
\includegraphics[width=1\linewidth]{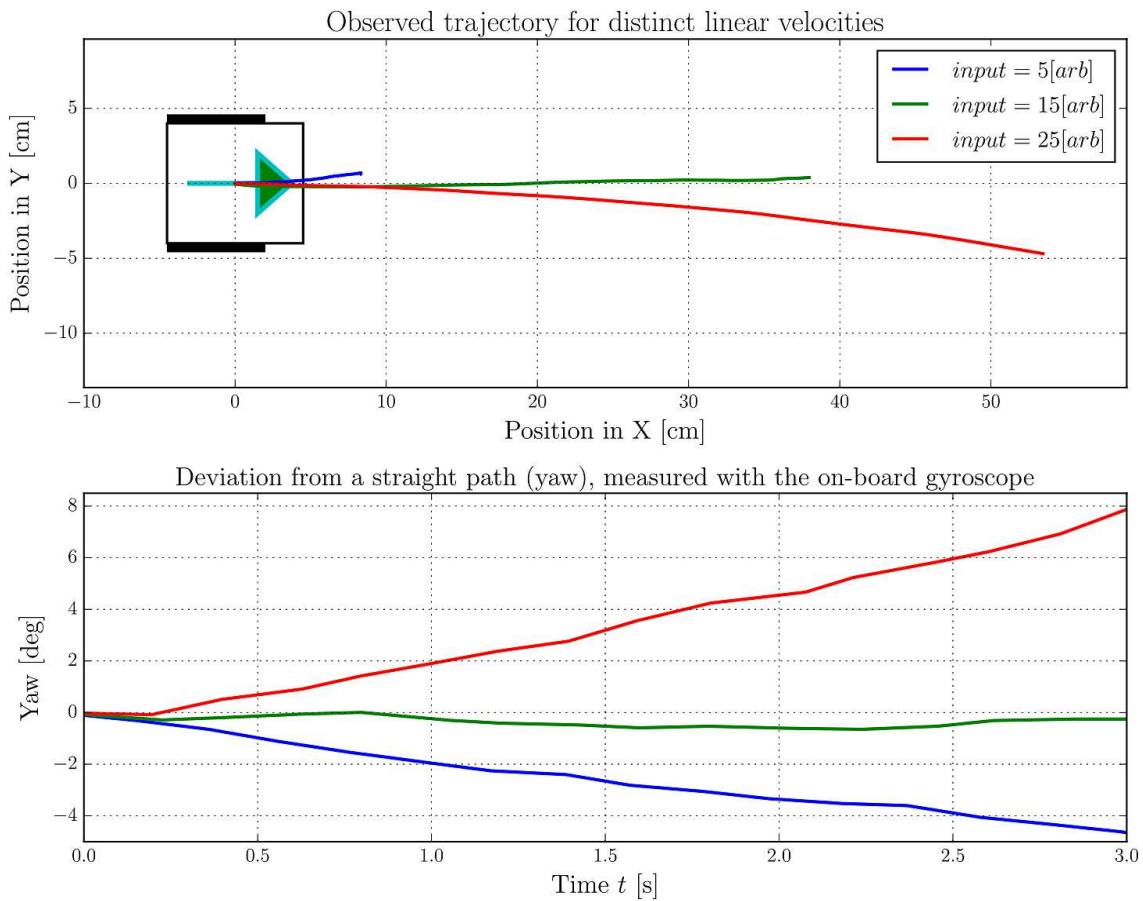}}
\caption[Initial experiments to analyse drift in the original controller]{\emph{Initial experiments to analyse drift in the original controller.} The robot was commanded to describe a straight path by setting both motors to the same velocity during 3 seconds.
The trajectory (upper panel) was recorded with a ceiling camera for three different velocities, while yaw (lower panel) was being logged using the on-board gyroscope.
It can be observed that drift is accurately tracked by the empirical yaw measurements.
The input speed setting is arbitrary since both continuous-rotation servomotors are commanded with the built-in $Servo.write([degrees])$ functions from the Arduino IDE, which do not have a direct mapping with real world velocities.
}
\label{fig:gnbotLinearDrift}
\end{figure}

The original controller on-board the GNBot was velocity-based and its input units were arbitrary -- it did not have an internal mapping between the motor's inputs and real-world units (i.e. $cm/s$). It was also lacking methods for basic calibration, in part due to the absence of any form of odometry. Altogether, these facts lead to large amounts of drift.
Far from being a constant that could have been compensated by basic calibration, drift varied with the velocity of the robots -not only in magnitude but also in direction-, so straightforward calibration was not an option (see Figure \ref{fig:gnbotLinearDrift}).
This project has tackled the problem by incorporating a gyroscope for odometry. The advantages of this approach in contrast with the use of wheel encoders were already discussed in Section \ref{sec:bg_section}.

The selected \emph{Inertial Measurement Unit} (IMU) is the MPU6050\footnote{\url{http://www.invensense.com/products/motion-tracking/6-axis/mpu-6050/}} chip, which is based on MEMS technology (Micro-Electro-Mechanical Systems). It is commonly available as an Arduino-compatible break-out board that conveniently fits on the $I^2C$ socket of the GNBot without any additional modifications (c.f. Figure \ref{fig:GNBotGyro}).

The gyroscope module has the same $I^2C$ pinout as the magnetometer that was originally present in the robot, but the routines that handle both modules are very different. In fact, the example code for the inertial sensor (a part of the \emph{$I^2C$ Device Library}\footnote{\url{http://www.i2cdevlib.com/devices/mpu6050}}) depends on the use of hardware \emph{interruptions}\footnote{\emph{Interrupt:} Signal to the processor indicating that some event needs immediate attention} rather than \emph{polling}\footnote{\emph{Polling:} As opposed to interruptions, polling actively samples the status of an external device}.
The use of interruptions was highly undesirable since the main processor needs to handle time sensitive tasks such as sensing and radio communications as well as motion control, so the gyroscope library was modified to use a polling scheme.

The MPU6050 incorporates a \emph{Digital Motion Processor} (DMP) that can be used to offload computations from the robot's main processor. These computations include the task of filtering and integrating the raw values from the on-chip acceleration sensors, in order to output ready to use \emph{yaw, pitch and roll}. In the GNBot controller the raw values are sampled and processed at $200Hz$ by the DMP co-processor.

The gyroscope module also has an internal FIFO\footnote{\emph{FIFO:} ``First In First Out'' queuing policy} buffer that allows for reliable high-frequency data readings (up to $200Hz$), though in this project it was deemed unnecessary since the sample rate is low ($<50Hz$). Once the IMU had been incorporated into the GNBot, it was possible to read the \emph{yaw, pitch and roll} of the robot back into the computer (see Figure \ref{fig:gnbotLinearDrift}).

At this point, a basic proportional yaw controller was implemented in order to demonstrate the viability of the system. The controller could now successfully correct the direction of the robot so it followed a straight path, even after external perturbations had been applied (i.e. rotating the robot or placing obstacles).

\begin{figure}[hbtp]
\centerline{\includegraphics[width=1\linewidth]{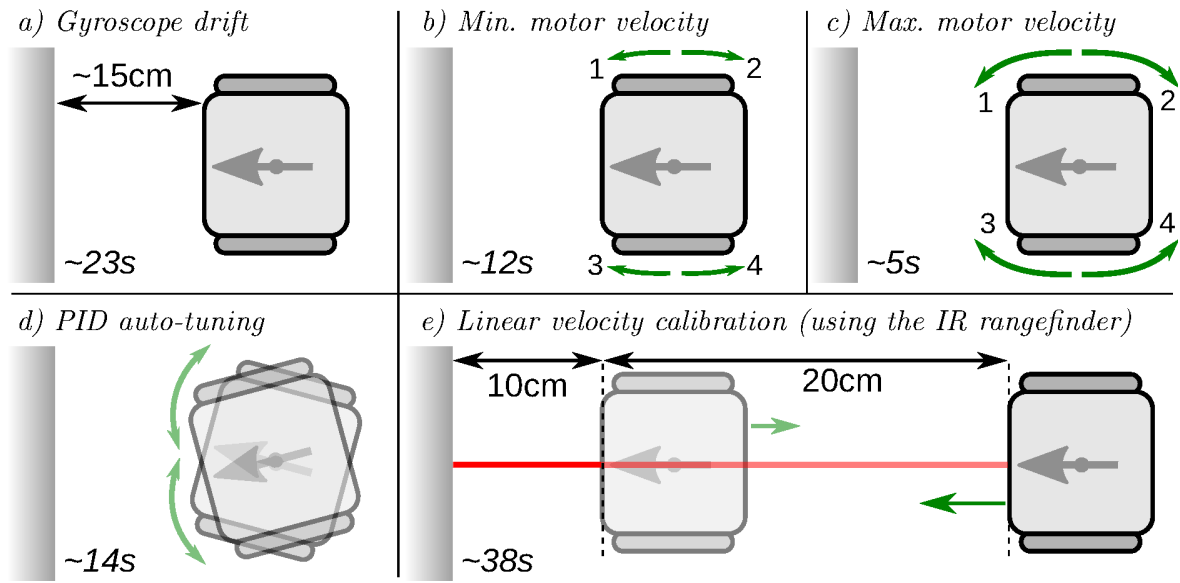}}
\caption[Full self-calibration procedure]{\emph{Full self-calibration procedure.} \textbf{\textit{(a)}} The robot is first placed in front of a wall, and remains static while the gyro drift is compensated, \textbf{\textit{(b)}} The minimum motion threshold is found by gradually increasing the input of each motor until a rotation is perceived \textbf{\textit{(c)}} The motors are then independently run at their maximum speed setting in order to record the corresponding real-world rotational velocity, \textbf{\textit{(d)}} Next, the yaw PID controller is heuristically calibrated by analysing the oscillations for different parameters, \textbf{\textit{(e)}} Finally, the robot uses the distance sensor to measure its approach velocity towards the wall, creating a map between the input values ($rad/s$) and the resulting real-world linear velocity ($cm/s$).\\
Technical details can be found in sections: $(a)$ \ref{sec:gyro_section}, $(b$ \& $c)$ \ref{sec:rotVel_section}, $(d)$ \ref{sec:PID_section}, $(e)$ \ref{sec:linearVel_section}.
}
\label{fig:full_calibration_procedure}
\end{figure}

With those results as a motivation, the design of a high-level self-calibrating motion controller was tackled. An overview of the resulting calibration routine is shown in Figure \ref{fig:full_calibration_procedure} -- Further sections will describe the technical details in greater depth.

\newpage
\section{Minimization of gyroscope drift} \label{sec:gyro_section}

Most electronic gyroscopes calculate rotational magnitudes by integrating measurements from rotational accelerometers over time.
Ideally this integration process would consistently yield the same outputs when a sequence of rotations is applied, but instead, rotational accelerometers often have inaccuracies (i.e. bias, sensitivity limits, variability with temperature, etc.) that cause \emph{drift} in the integration process.
This would imply for instance that the yaw measurement of a static robot would erroneously vary over time.

\begin{figure}[hbtp]
\centerline{
\includegraphics[width=0.9\linewidth]{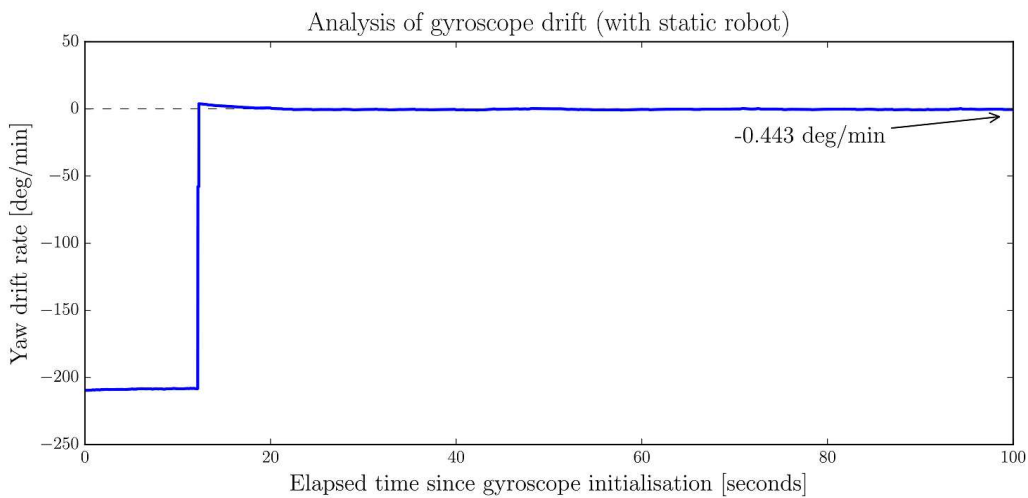}}
\caption[Analysis of gyroscope drift (with static robot)]{\emph{Analysis of gyroscope drift (with static robot).} The gyroscope has an initial transient that must be respected, during which the internal DMP (Digital Motion Processor) performs self-calibration to account for drift. The module is fully calibrated after 23 seconds of the robot being static, when drift becomes inferior to 0.5 deg/min (step A in Fig. \ref{fig:full_calibration_procedure}). It must be noted though that the magnitude of this drift will not be constant, but is instead greatly affected by the robot's motions.
}
\label{fig:gyro_drift}
\end{figure}

Gyroscope drift is often minimized with either a proper calibration of the accelerometer bias, with signal filtering or by pausing the integration process while no motion is detected. Fortunately, the DMP co-processor of the MPU6050 already implements these features very efficiently (see Fig. \ref{fig:gyro_drift}).

In order to measure yaw drift rate (whose units are \emph{degrees per minute}) it was necessary to differentiate the yaw measurements provided by the gyroscope (\emph{radian} units). This was achieved by accurately timing distinct yaw measurements with the function \emph{millis()} from Arduino\footnote{\url{https://www.arduino.cc/en/Reference/Millis}}.
\begin{equation}
drift\_rate=\frac{yaw(t_2)-yaw(t_1)}{t_2-t_1}\cdot\frac{180 deg}{\pi rad}\cdot\frac{1 min}{60 s}deg/min
\end{equation}

\vspace{-5mm}

\section{Calibration of the motors} \label{sec:rotVel_section}

This section tackles the calibration of the continuous-rotation servomotors, which are actuators that generate a rotational velocity that is proportional to a PWM input signal (Pulse Width Modulation).
The relationship between the input and output magnitudes is arbitrary, so a real-world mapping needs to be learned first.

To better understand the calibration process it helps to be familiar with the following notation:

\begin{figure}[hbtp]
\centerline{
\includegraphics[width=0.33\linewidth]{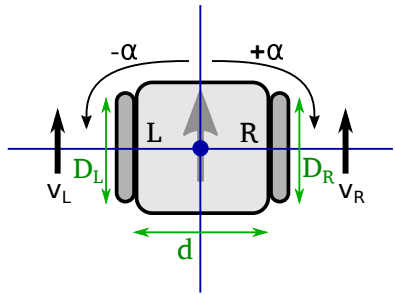}}
\caption[Diagram of the robot's top view, displaying the notation used]{\emph{Diagram of the robot's top view, displaying the notation used.}
The grey arrow points towards the front of the robot.
$D_L$ and $D_R$ are the wheel diameters, and $d$ is their separation. $V_L$ and $V_R$ are the contributions of each motor to the robot's linear velocity, and $\alpha$ is the yaw measurement (positive for clockwise rotations).
}
\label{fig:robot_diagram_notation}
\end{figure}

Most mobile robotic platforms have encoders that simplify the calibration process by measuring the real-world rotational velocities of the wheels. In the case of the GNBot there are no wheel encoders, so a different approach was needed.

\begin{figure}[hbtp]
\centerline{
\includegraphics[width=0.75\linewidth]{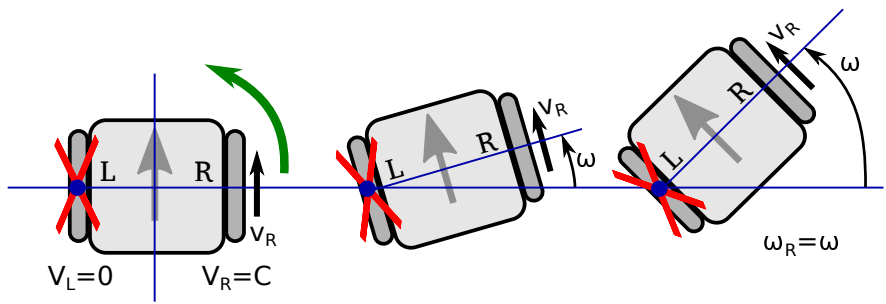}}
\caption[Method employed to measure $\omega_R$, the contribution of the right motor to the robot's rotational velocity $\omega$.]{\emph{Method employed to measure $\omega_R$, the contribution of the right motor to the robot's rotational velocity $\omega$.}
A constant velocity is first applied to the right wheel ($V_R=C$) while the left wheel is fixed ($V_L=0$).
This results in a rotation of the robot around the ground-contact point of the left wheel. The gyroscope is then used to measure the rotational velocity $\omega_R=\omega$.
Afterwards the same process is repeated for the left motor.
This method allows to separately evaluate the response curves of each motor without requiring wheel encoders.
}
\label{fig:speed_calib_diagram}
\end{figure}

The implemented method can measure the response curve of each motor independently by using only the gyroscope. This process is described in Figure \ref{fig:speed_calib_diagram}.

\begin{figure}[hbtp]
\centerline{
\includegraphics[width=1.15\linewidth]{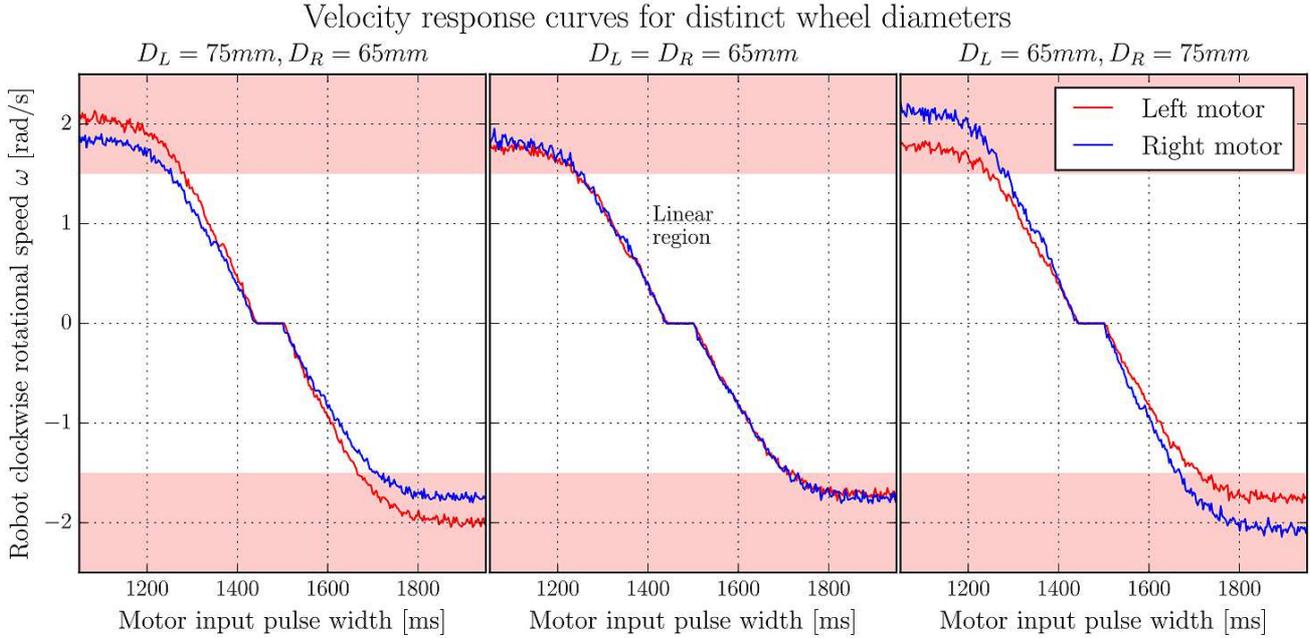}}
\caption[Measured motor response curves]{\emph{Measured motor response curves.}
The panels show the response curves of the left (red) and right (blue) motors for different wheel sizes.
The input of the continuous-rotation servomotors is PWM (Pulse Width Modulation), and the output ($\omega$) is the rotational rate of the robot around its vertical axis (the setup is described in Fig. \ref{fig:speed_calib_diagram}).
Each trial was run for a single motor at a time, and the sweep took around 30 seconds per motor.
Three regions can be appreciated: a flat \emph{dead-zone} ($\omega=0rad/s$), a linear response region ($0 < \omega < 1.5rad/s$), and saturation ($\omega \geq 1.5rad/s$, which has been highlighted in red).
For the same wheel size -middle panel- the L/R response curves are almost identical.
Robots with larger left or right wheels -left \& right panels- instead have visibly different L/R response curves.
The calibration routine implemented in this project accounts for these differences.
}
\label{fig:velocity_response_curve}
\end{figure}

The angular velocity $\omega$, directly measured by the gyroscope, is coupled with the rotational speed of both motors with constants $K_L$ \& $K_R$, that account for $D_L$, $D_R$ (wheel diameters), and $d$ (distance between wheels).
Using this technique it was possible to record the response curve of each motor by independently performing a sweep over the full range of velocity inputs.
The result is shown in Figure \ref{fig:velocity_response_curve}.

Self-calibration of these curves has been implemented by means of linear fitting between the minimum and maximum velocities within the linear region. In first place, the robot measures the \emph{dead-zone} of a motor (the minimum motion threshold) by gradually increasing its input velocity until a rotation is perceived. Then, the maximum velocity is measured by running the motor at the maximum speed setting within the linear range (PWM$\approx$1300 or PWM$\approx$1650).
These measurements are performed for the forward and backward directions of each motor (steps B \& C in Fig. \ref{fig:full_calibration_procedure}).

\begin{figure}[hbtp]
\centerline{
\includegraphics[width=0.7\linewidth]{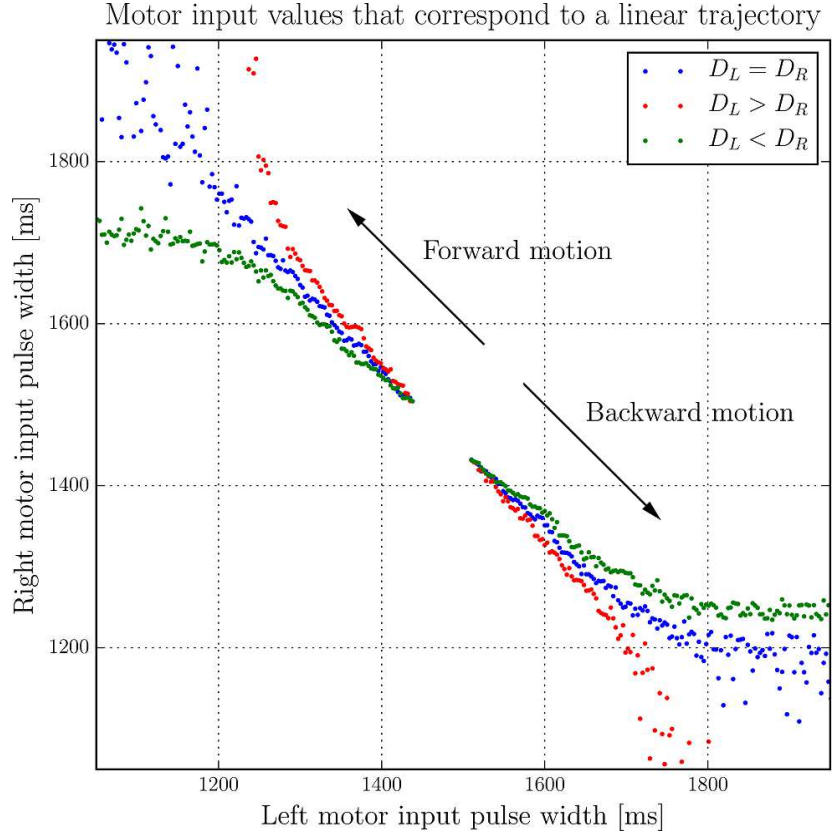}}
\caption[Region of motor input values that correspond to a linear trajectory]{\emph{Region of motor input values that correspond to a linear trajectory.}
The red and green data points are from robots with larger left or right wheels respectively, while the blue data points are from a robot with identical wheels. It can be observed that each of these curves has a different steepness: a greater slope indicates that the right motor needs to rotate faster in order to account for a larger left wheel, and vice-versa.
}
\label{fig:linear_trajectory_PWM_mapping}
\end{figure}

The left and right motors contribute to the robot's global angular velocity $\omega$ in an additive manner ($\omega=\omega_L+\omega_R$, i.e. setting $\omega_L=0.5rad/s$ and $\omega_R=0.5rad/s$ yields $\omega=1.0rad/s$).
Knowing this fact, and using the data from Figure \ref{fig:velocity_response_curve}, it was possible to determine the input region that corresponds to a linear motion. This is when both rotational contributions cancel out so the robot does not rotate: $\omega=\omega_L+\omega_R=0rad/s \rightarrow \omega_L=-\omega_R$.
This region is represented in Figure \ref{fig:linear_trajectory_PWM_mapping}.

\section{PID auto-tuning using the Ziegler-Nichols method} \label{sec:PID_section}

At this point a basic motor controller had been implemented, so the rotational rate ($rad/s$) of the robot could be accurately commanded. Closed-loop yaw control was now possible with the implementation of a PID regulator (Proportional-Integral-Derivative) that combines the  motor controller with real-time gyroscope measurements (Eqn. 2.2). The error measure is calculated as $e(t)=yaw_{target}-yaw(t)$.
\begin{equation}
output\_rotation\_rate(t) = K_p e(t) + K_i \int_{0}^{t}e(\tau)d\tau + K_d \frac{d}{dt}e\ \ [rad/s]
\end{equation}

\begin{figure}[hbtp]
\centerline{\includegraphics[width=0.97\linewidth]{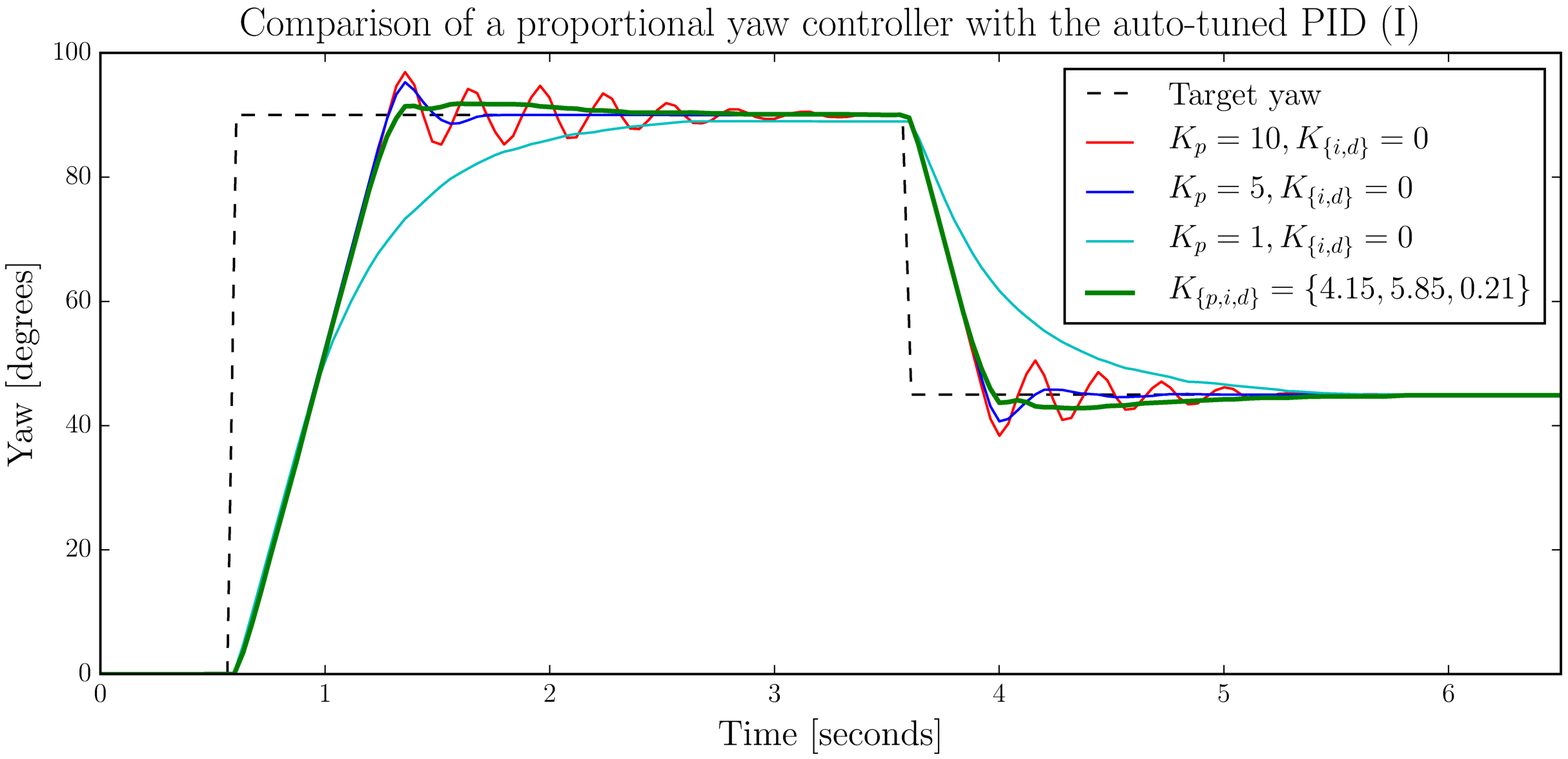}}
\vspace{-3mm}
\centerline{\includegraphics[width=0.99\linewidth]{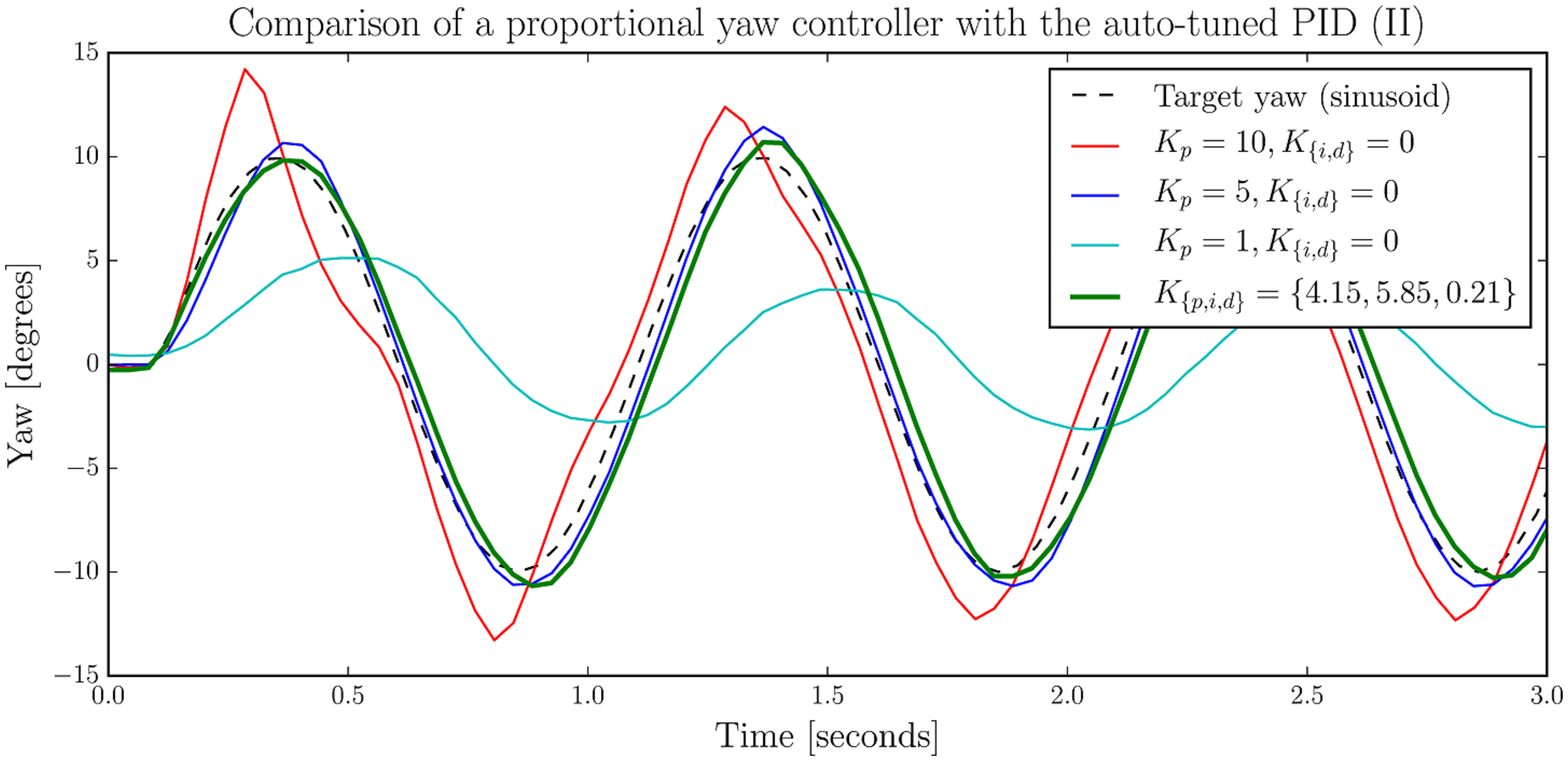}}
\caption[Response of the yaw controller to a step function and to a sinusoid]{\emph{Response of the yaw controller to a step function (upper panel) and to a sinusoid (lower panel).}
The controller was commanded with time-varying yaw targets while the transitions were recorded using the gyroscope.
It can be appreciated that a proportional controller either is too slow (cyan curve), has over-shoot (blue curve) or even ripple (red curve).
The auto-tuned PID (green curve) does have some overshoot as well, but it is a much better approximation to the ideal response.
}
\label{fig:P_controller_vs_autotuned_PID}
\end{figure}

A PID controller is based on user-tunable gains ($K_p$, $K_i$ \& $K_d$) that, when properly adjusted, can minimize the overshoot of the transient response; in the case of the GNBot, the PID is the responsible of achieving fast and accurate changes in yaw. This is represented in Figure \ref{fig:P_controller_vs_autotuned_PID}.

Rather than manually specifying a fixed parameter set for the PID, this project tackled the self-calibration of the controller with the popular technique proposed by Ziegler and Nichols\cite{ziegler1942optimum}. This heuristic tuning method is performed by setting the three gains ($K_p$, $K_i$ \& $K_d$) to zero, and then increasing the proportional term $K_p$ until it reaches the ultimate gain $K_u$ at which the controller presents self-sustained oscillations of period $T_u$.
Ziegler and Nichols (ZN) then provide with the mapping between $K_u$ \& $T_u$ and the three gains of the PID controller.
In this project the modified tuning values proposed by Tyreus and Luyben\cite{luyben1997essentials} were used instead as they increased the robustness of the controller. Table 2.1 compares the gain specifications of both tuning methods.

\begin{table}[hbtp]
\centering
\begin{tabular}{l|l|l|l|}
\cline{2-4}
                                                              & \cellcolor[HTML]{EFEFEF}$K_p$ & \cellcolor[HTML]{EFEFEF}$t_I$ & \cellcolor[HTML]{EFEFEF}$t_D$ \\ \hline
\multicolumn{1}{|l|}{\cellcolor[HTML]{EFEFEF}Ziegler-Nichols (ZN)} & $K_u/1.7$                     & $T_u/2$                       & $T_u/8$                       \\ \hline
\multicolumn{1}{|l|}{\cellcolor[HTML]{EFEFEF}Tyreus-Luyben (TLC)}   & $K_u/2.2$                     & $2.2T_u$                      & $T_u/6.3$                     \\ \hline
\end{tabular}
\caption{\emph{Prescribed PID gain values for the Ziegler-Nichols (ZN) \& Tyreus-Luyben (TLC) heuristic tuning methods.}
Compared to the original values proposed by Ziegler \& Nichols, the TLC rules tend to reduce the oscillatory effects and improve the robustness of the yaw controller.
The $t_I$ and $t_D$ values were converted into the standard PID form using $K_i = K_p/T_i$ and $K_d = K_pT_d$.
}
\label{my-label}
\end{table}

\begin{figure}[hbtp]
\centerline{
\includegraphics[width=1.2\linewidth]{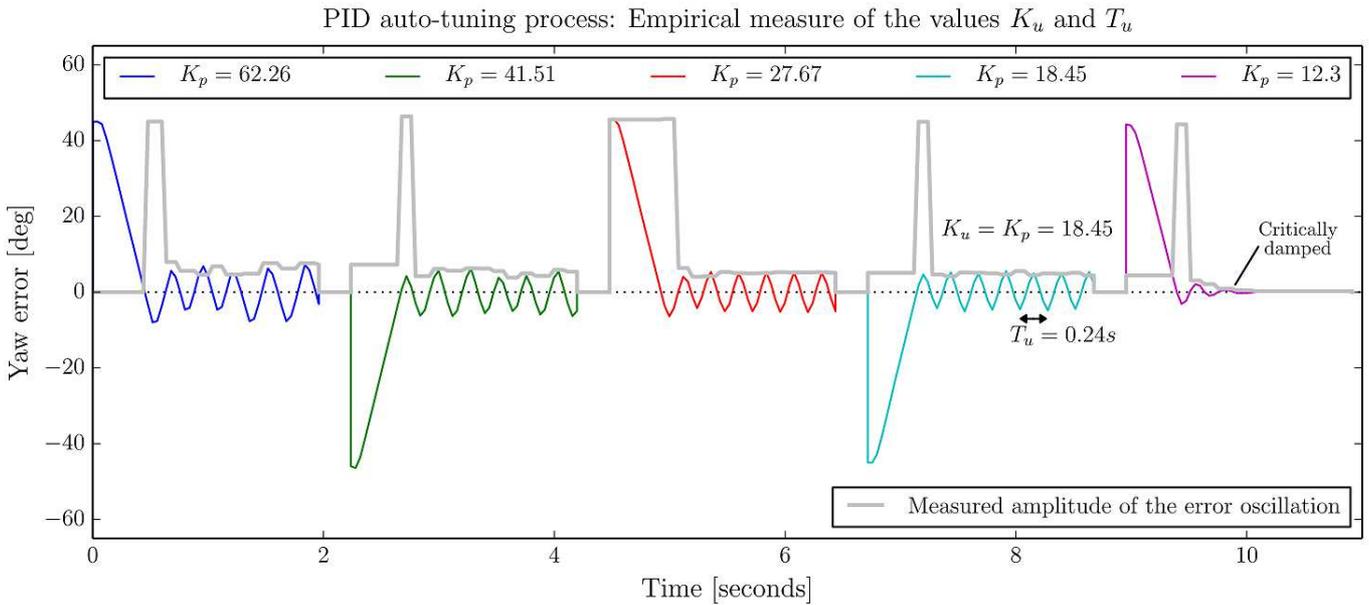}}
\caption[PID auto-tuning process: Empirical measure of the values $K_u$ \& $T_u$]{\emph{PID auto-tuning process: Empirical measure of the values $K_u$ \& $T_u$.}
The tuning is done as follows: In first place, $K_i$ \& $K_d$ are set to zero, and $K_p$ is set to a very large value that ensures oscillation (the value is such that the maximum motor speed is applied at 1 degree of error). Next, the proportional controller is perturbed by setting a 45 degree angle as the target yaw. Then the algorithm checks whether the oscillation is self-sustained, in which case it updates $K_p=K_p/1.5$. This process is repeated until the oscillations become attenuated; at that point $K_u$ \& $T_u$ are finally registered (step D in Fig. \ref{fig:full_calibration_procedure}).
}
\label{fig:PID_auto_tuning}
\end{figure}

In order to provide a consistent PID calibration, the parameters $K_u$ -ultimate gain- and $T_u$ -oscillation period- needed to be measured accurately and reliably. An automated method was implemented for this purpose (see Figure \ref{fig:PID_auto_tuning}).

\section{Mapping rotational velocities with linear motions} \label{sec:linearVel_section}

Using the basic motion controller it was now possible to command the robot to perform linear trajectories quite accurately; but the inputs units were still rotational magnitudes related to yaw ($\omega_L$ \& $\omega_R$ $[rad/s]$). The next step was to create a mapping between these inputs and the real-world linear velocities of the robot (i.e. $cm/s$).

In first place it was necessary to have a method that could accurately measure the robot's motions.
External tracking systems (i.e. VICON\footnote{\url{http://www.vicon.com/products/camera-systems}} or a ceiling camera) are among the best solutions available for this purpose, but they were deemed too sophisticated to be incorporated into the basic calibration routine. These methods were employed for the final evaluation of the motion controller instead (Chapter \ref{sec:eval_chap}).

\begin{figure}[hbtp]
\centerline{
\includegraphics[width=1.1\linewidth]{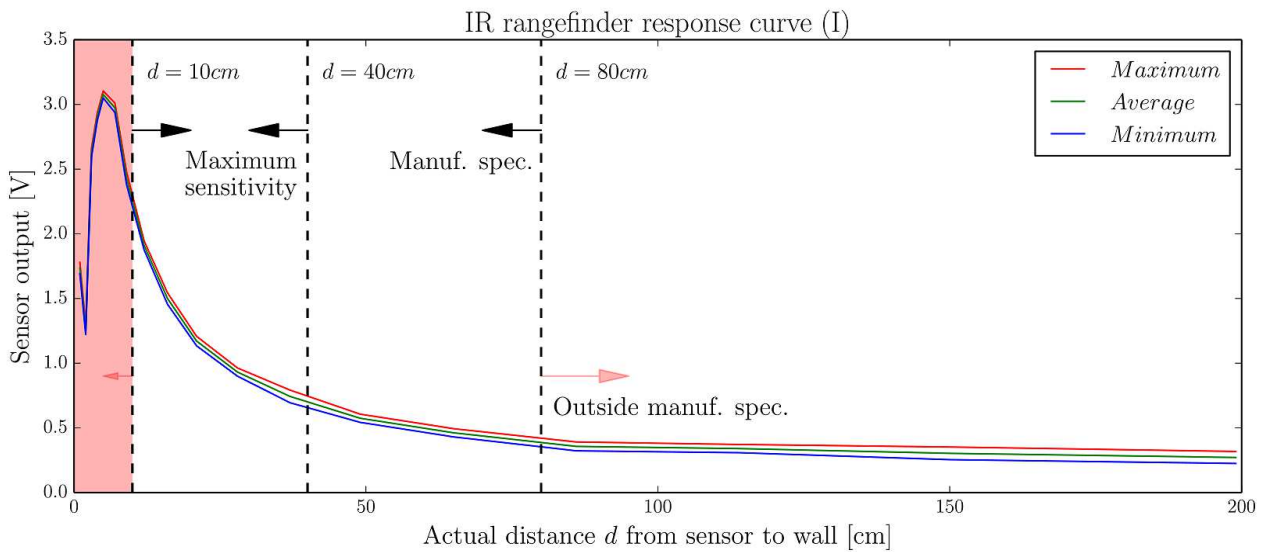}}
\caption[Analysis of the response curve of the infrared distance sensor]{\emph{Analysis of the response curve of the infrared distance sensor.}
The figure displays the output voltage of the \emph{Sharp GP2Y0A21YK0F} rangefinder and its variability with the distance between the sensor and a wall. Manufacturer specifications rate this module for distance measurements in the range $10cm<d<80cm$, and indeed it can be appreciated that the response outside that region is either flat ($d>80cm$) or nonlinear ($d<10cm$, highlighted in red).
}
\label{fig:IR_sensor_response_curve_B}
\end{figure}

Rather than requiring an external tracking system, the problem of measuring the robot's real-world velocity has been tackled with the use of the distance sensor on-board the GNBot.
The sensor is the \emph{Sharp GP2Y0A21YK0F}\footnote{\url{http://www.sharpsma.com/optoelectronics/sensors/distance-measuring-sensors/GP2Y0A21YK0F}} infrared rangefinder, and it has the voltage response curve analysed in Figure \ref{fig:IR_sensor_response_curve_B}.

The approach undertaken towards the calibration of linear motions was the use of the IR distance sensor on-board the GNBot to achieve accurate real-world velocity measurements. Theoretically, the robot would be driven towards a wall at a constant speed, and then the robot's velocity could be calculated by differentiating distance measurements over time.

\begin{figure}[hbtp]
\centerline{
\includegraphics[width=1.1\linewidth]{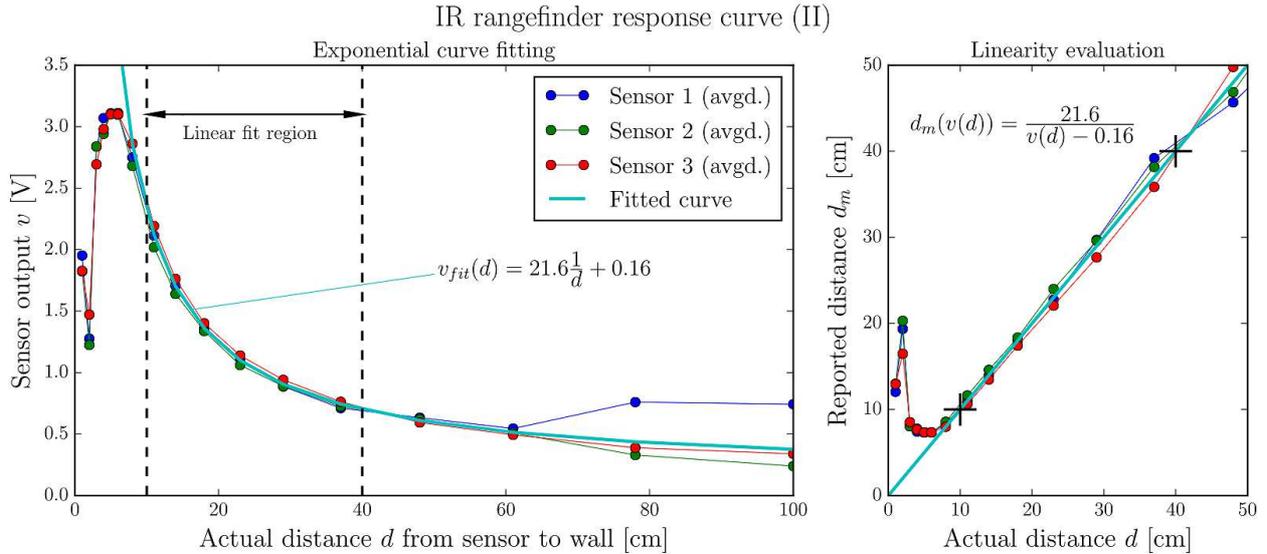}}
\caption[Linearisation of the IR rangefinder sensor response]{\emph{Linearisation of the IR rangefinder sensor response.}
The left panel shows the response curves that were recorded for three different GP2Y0A21YK0F distance sensors (red, green and blue curves).
These curves were fitted with an exponential decay (cyan curve) in the range of highest sensitivity ($10cm<d<40cm$).
The right panel evaluates the accuracy of the fitted model, showing a maximum theoretical deviation of $\approx3cm$.
}
\label{fig:IR_sensor_response_curve}
\end{figure}

In order to achieve a fair degree of resolution in these measurements, the output of the IR sensor needed to be calibrated first by means of linearisation. The method is introduced in Figure \ref{fig:IR_sensor_response_curve}.

The fitting function in Figure \ref{fig:IR_sensor_response_curve} is an inverse function with two fitting parameters $K$=$21.6V/cm$ \& $C$=$0.16V$ that were calculated using two data points located in the limits of the highest sensitivity range ($d_1$=$10cm$ \& $d_2$=$40cm$), using the following equations:
\begin{equation}
v_{fit}(d)=K\cdot\frac{1}{d}+C \ \ \ \rightarrow\ \ \ 
K = d_1\frac{v(d_1)-v(d_2)}{1-\frac{d_1}{d_2}} \ \ \ \ \ \ 
C = v(d_2)-\frac{K}{d_2}
\end{equation}
The calibration of the IR distance sensor is \emph{not} a part of the self-calibration routine since it requires user interaction; instead, the fitting curve can be generalised to every robot that uses the GP2Y0A21YK0F sensor model.

\begin{figure}[hbtp]
\centerline{
\includegraphics[width=1.05\linewidth]{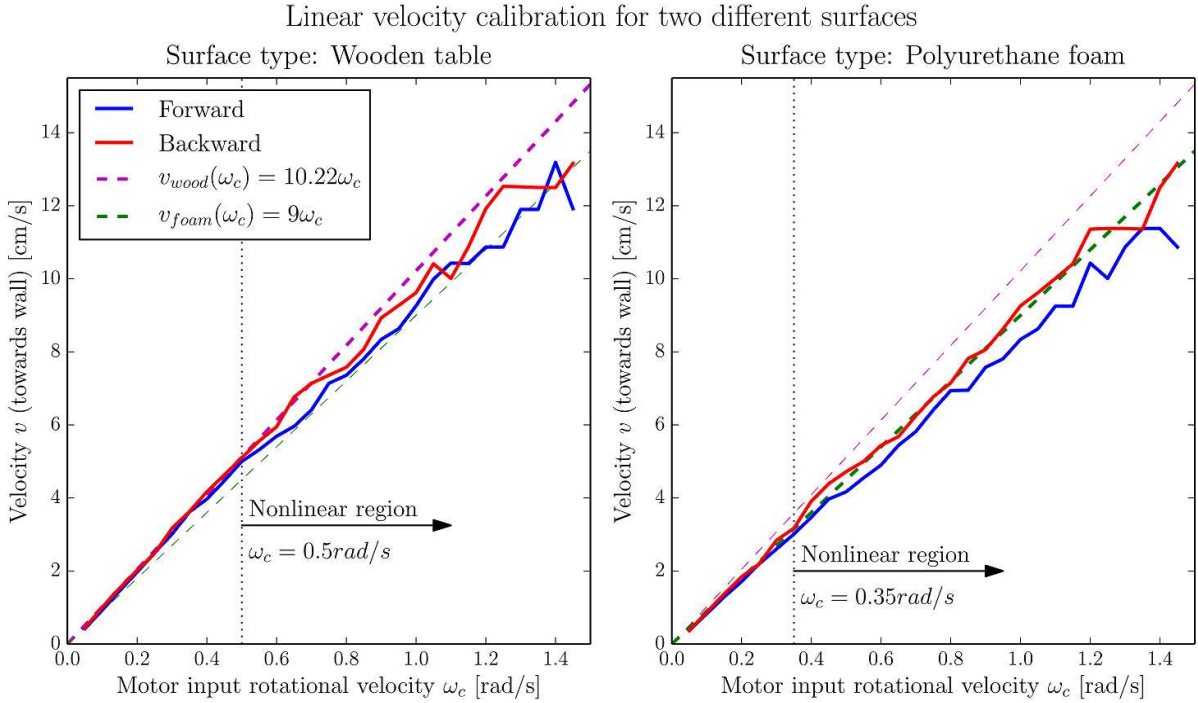}}
\caption[Linear velocity calibration for two different ground surfaces]{\emph{Linear velocity calibration for two different ground surfaces.}
The panels show how the real-world velocity of the robot varies over its input range, for both backward and forward motions (red and blue curves) and for two different ground surfaces (left and right panels).
The green and magenta dashed lines represent a linear fitting of the velocity response of each material.
It can be appreciated that the speed of the robot over foam (right panel) is lower for the same input values; this is due to friction effects.
Friction also causes an asymmetry in the velocity responses for backward and forward motions; this is because the robot is not perfectly symmetric.
The nonlinear region is a cause of the PID operating the motors outside their linear range (c.f. Fig. \ref{fig:velocity_response_curve});
such effect could be minimized with velocity-dependent PID gains.
}
\label{fig:linear_velocity_calibration_curves}
\end{figure}

This newly-calibrated distance sensing functionality was verified to work within the theoretically-calculated tolerances for different wall surfaces.
Having this infrastructure in place, the problem of linear velocity calibration could finally be tackled.
For this purpose, the GNBot was commanded to move towards a wall at different velocities while yaw was corrected by the closed-loop PID controller. The real-world linear approach velocity ($v\ cm/s$) was then recorded for multiple input velocity settings ($\omega_c\ rad/s$), yielding the results shown in Figure \ref{fig:linear_velocity_calibration_curves}.

Self-calibration was then implemented by means of a linear fitting that measures the real-world velocity at the highest velocity setting within the linear region. This measurement is performed four times and then averaged (step E in Fig. \ref{fig:full_calibration_procedure}), being the last step of the self-calibration process.
Each of the calibration parameters are then stored into the non-volatile EEPROM memory built into the Arduino, so the tuning process is not required every time the robots are powered on.

Finally, high-level distance-based control was implemented by integrating over time the current velocity setting; this way the robot can stop whenever a specified distance has been reached. This is a very simple method that does not account for transients in velocity.
Arc functionality was implemented by interpolating the target yaw throughout a distance-based motion; specifying different initial/final yaw angles effectively converts a linear segment into an arc.

%%%%%%%%%%%%%%%%%%%%%%%%%%%%%%%%%%%%
\chapter{Evaluation of the motion controller} \label{sec:eval_chap}

The purpose of the controller described in Chapter \ref{sec:calib_chap} is to provide accurate high-level motion control for the GNBot robots.
This chapter evaluates the real-world performance of this controller.

\begin{figure}[hbtp]
\centerline{\includegraphics[width=0.7\linewidth]{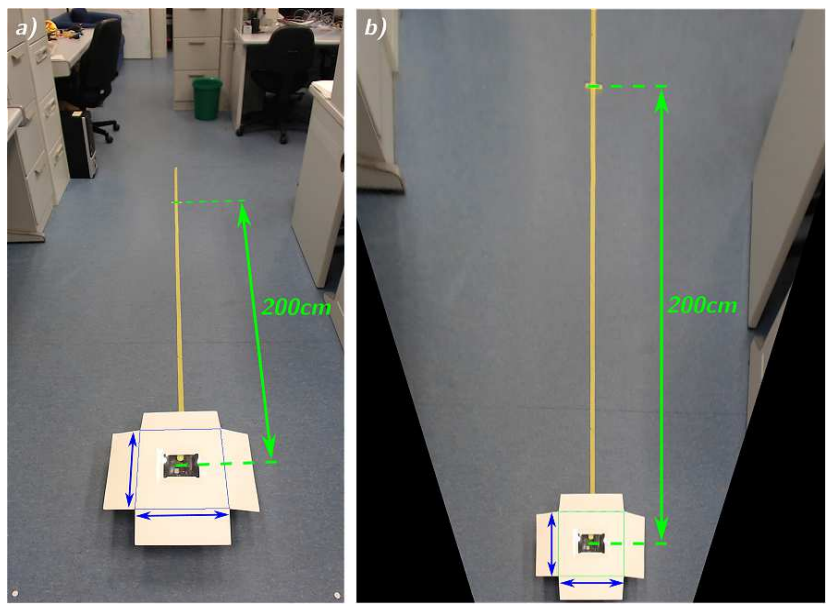}}
\caption[Perspective correction for accurate video tracking]{\emph{Perspective correction for accurate video tracking.}
The video tracking system was calibrated using perspective correction with a square of known dimensions. This allows for a 1:1 mapping between image units and real-world distances in the XY plane. 
}
\label{fig:perspective_correction}
\end{figure}

\begin{figure}[hbtp]
\centerline{\includegraphics[width=0.9\linewidth]{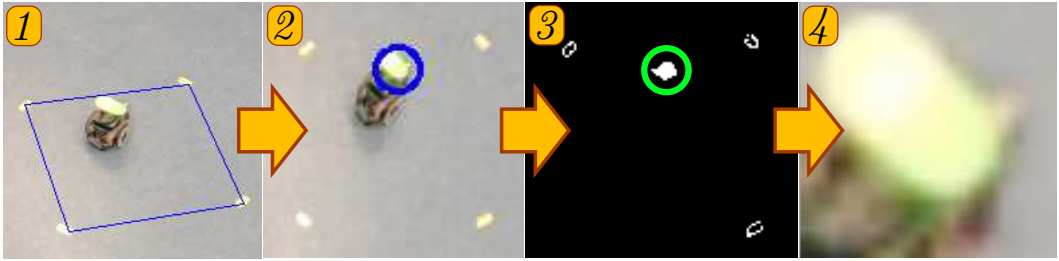}}
\caption[How color markers are tracked with the OpenCV library]{\emph{How color markers are tracked with the OpenCV library.}
In first place the reference points for perspective correction need to be specified by using the cursor \emph{(1)}. Next, the marker color is selected \emph{(2)}, and then a threshold is applied \emph{(3)}. The algorithm finally looks for the largest blob, which yields the coordinates of the robot \emph{(4)}. From \cite{GarciaSaura14}.
}
\label{fig:opencv_tracking_steps}
\end{figure}

In first place, a series of experiments were designed in order to be able to measure the different \emph{motion drift} characteristics. For instance, the robots would be commanded to perform squares and circles of known dimensions while external video tracking provided unbiased measurements of the actual trajectories. 
The tracking process is described in Figures \ref{fig:perspective_correction} and \ref{fig:opencv_tracking_steps}.

Using this tracking system it was possible to record the trajectories of each robot with a very high resolution. Next sections describes the experiments that were performed.

\vspace{-5mm}

\section{The effect of gyroscope drift} \label{sec:sect31}

\vspace{-5mm}

\begin{figure}[hbtp]
\centerline{\includegraphics[width=0.55\linewidth]{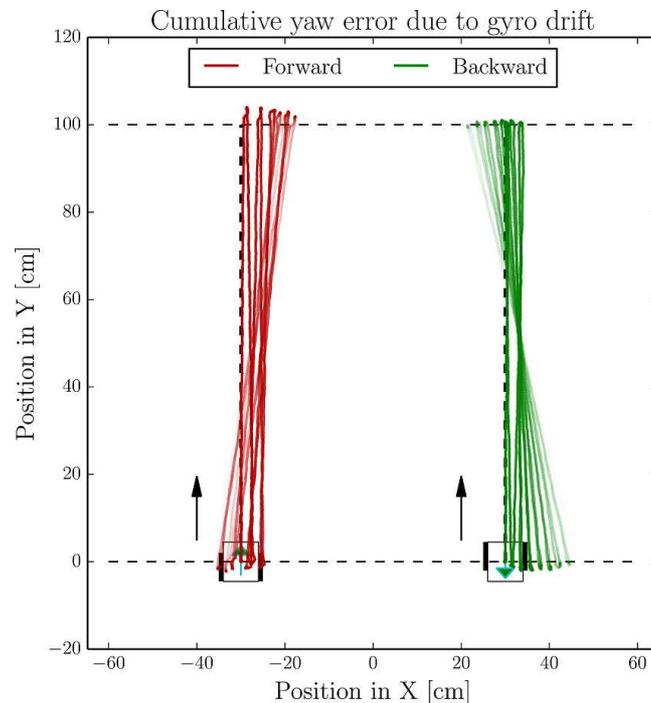}}
\vspace{-2.5mm}
\caption[Cumulative yaw error due to gyroscope drift]{\emph{Cumulative yaw error due to gyroscope drift.}
In this experiment, the robot was commanded to repeat a linear trajectory multiple times (alternating the target yaw between $0^{\circ}$ and $180^{\circ}$) for both forward motion (red path) and backward motion (green path)
at a velocity setting of $5cm/s$.
Using this data, the rotational drift of the controller has been calculated to be in the order of $2\ deg/min$ in a moving robot.
}
\label{fig:cumulative_yaw_error}
\end{figure}

Though gyroscope drift is always compensated upon the initialization of the GNBots, such compensation is far from ideal, and as a consequence the robot's motions present yaw drift over time. This effect is analysed in Figure \ref{fig:cumulative_yaw_error}.

\section{Analysis of linear positioning accuracy} \label{sec:sect32}

\begin{figure}[hbtp]
\centerline{\includegraphics[width=1\linewidth]{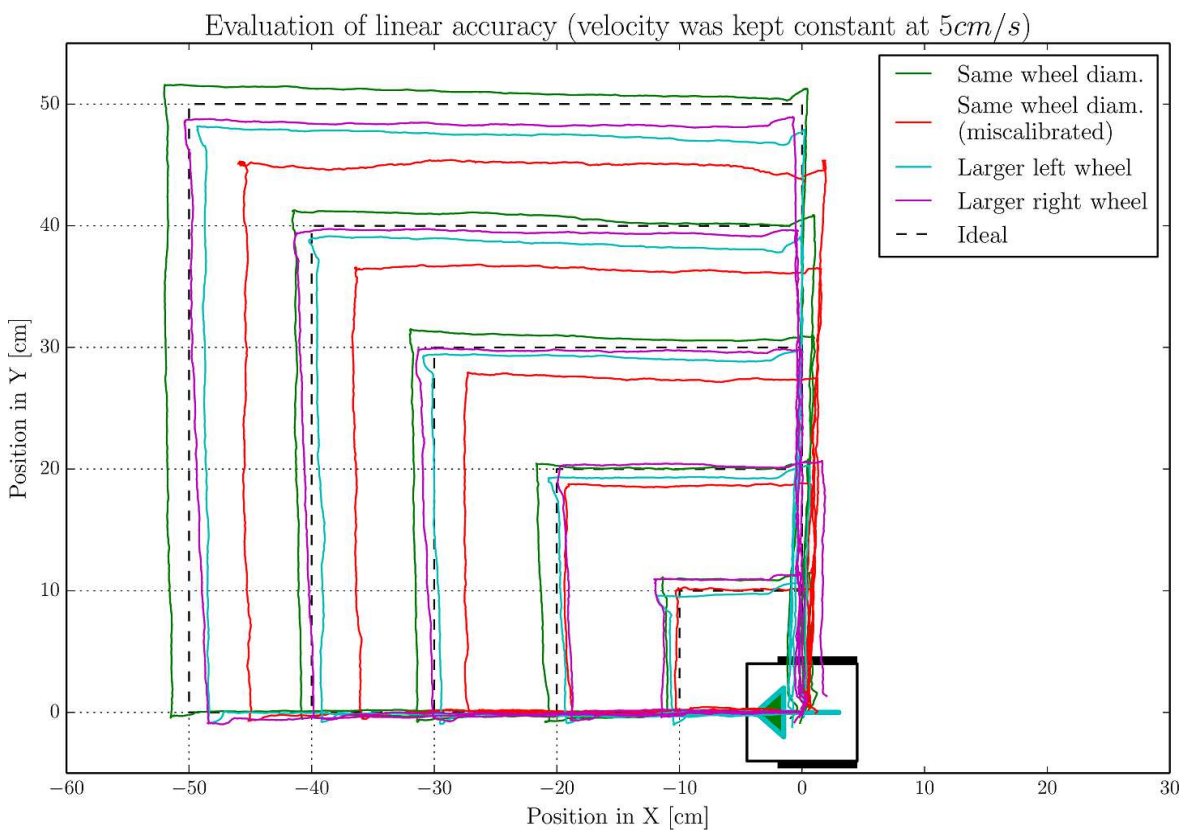}}
\caption[Evaluation of the linear accuracy of the distance-based controller]{\emph{Evaluation of the linear accuracy of the distance-based controller.}
A GNBot robot (lower-right corner) was commanded to describe square trajectories of different sizes while its path was recorded using the vision-based method described in Figures \ref{fig:perspective_correction} \& \ref{fig:opencv_tracking_steps}.
The red path corresponds to a purposely-miscalibrated robot (it was calibrated with a different set of wheels).
The cyan and magenta paths correspond to robots with different wheel sizes that were properly self-calibrated.
Finally, the green path corresponds to a self-calibrated robot with same wheel diameters.
Linear drift has been calculated to be around $3cm/m$. This was done by scaling up the maximum deviation from the lower-left corner of the largest square trajectory -- about $1.5\ cm$ for the calibrated robots.
It can be appreciated that the shape of the squares is correctly maintained thanks to the gyroscope -which is always calibrated upon start up-, and the square dimensions only fall out of the specification for the miscalibrated robot (red path).
}
\label{fig:square_distance_comparison}
\end{figure}

Next, the general performance of the distance-based controller was evaluated by first commanding the robots to perform squares and circles and then comparing the results against ground-truth dimensions.
Square trajectories were implemented with straight line segments interleaved with in-place rotations of $90^{\circ}$. This functionality yielded the results shown in Figure \ref{fig:square_distance_comparison}.

In order to simplify the video tracking process, measurements were conducted in one single continuous experiment for each robot. In the case of square trajectories, the GNBot would perform the different square sizes one after another -- without an intermediate re-positioning of the robot.
This generated an undesired drift in the starting position of each trial, which made manual alignment necessary.
The post processing consisted in a affine translation from the starting points to the origin ($x,y$=$0,0$) and posterior re-setting of the initial yaw angle by means of a global rotation.
This process effectively removed the effects of yaw drift between each trial (the effect had already been analysed in Figure \ref{fig:cumulative_yaw_error}).

\begin{figure}[hbtp]
\centerline{\includegraphics[width=1\linewidth]{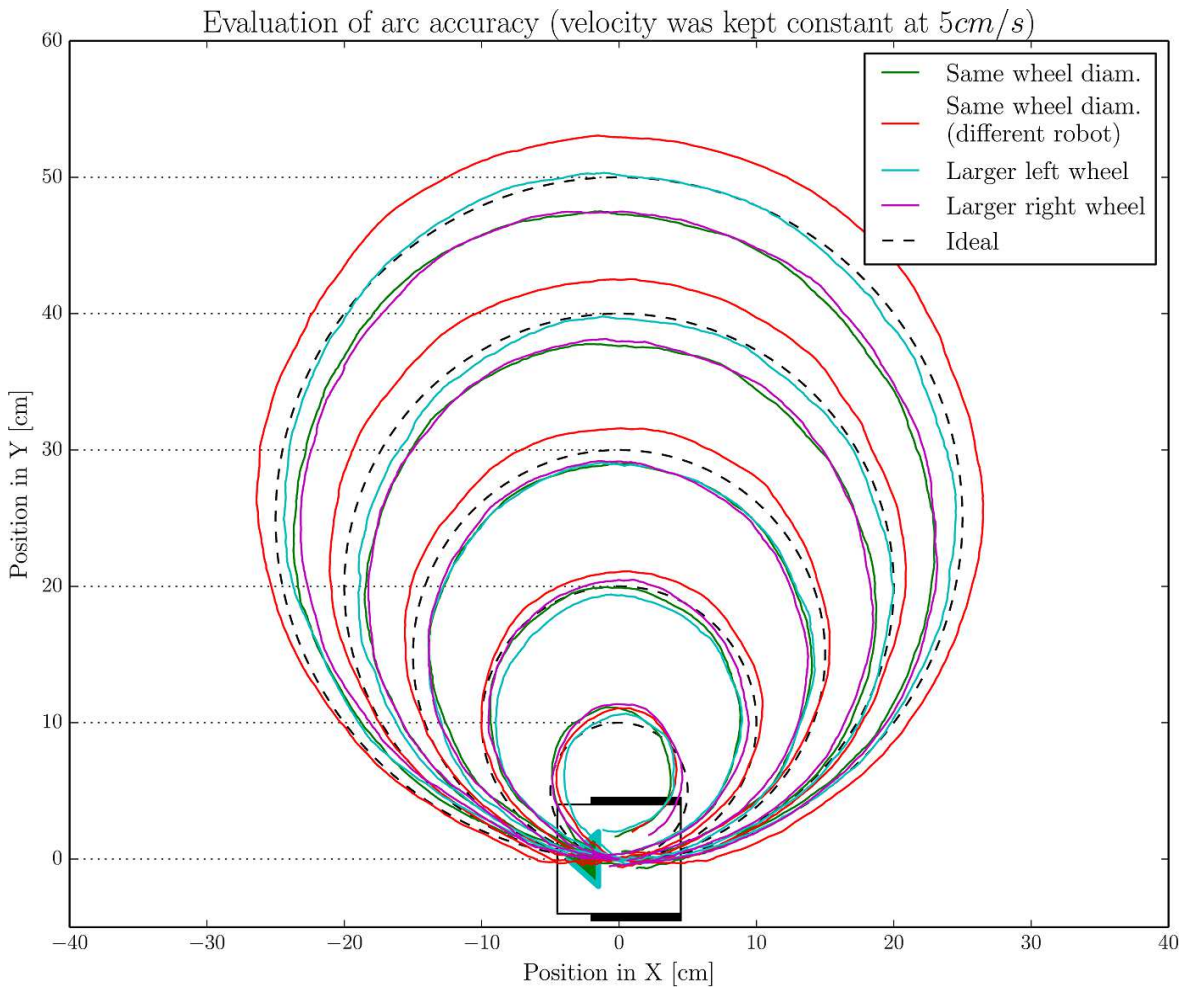}}
\caption[Evaluation of the accuracy of arcs performed by the controller]{\emph{Evaluation of the accuracy of arcs performed by the controller.}
A GNBot robot (lower-right corner) was commanded to describe circular trajectories of different sizes while its path was recorded.
The cyan and magenta paths correspond to robots with different wheel sizes that were properly self-calibrated.
The green path corresponds to a self-calibrated robot with same wheel diameters.
Finally, the red path corresponds to a different robot that was also self-calibrated.
The maximum drift from ground-truth diameters is around $8cm/m$, though it can be appreciated that the circular shape is correctly maintained thanks to the gyroscope
(which is always calibrated upon start-up).

}
\label{fig:circle_distance_comparison}
\end{figure}

Circular motions were implemented with the arc functionality, by setting the initial and final arc angles to $0^{\circ}$ and $360^{\circ}$. Path length was calculated as $l$=$2 \pi r$.
This approach yielded the results shown in Figure \ref{fig:circle_distance_comparison}.

\newpage
\section{How velocity affects the controller} \label{sec:sect33}

\begin{figure}[hbtp]
\centerline{\includegraphics[width=0.95\linewidth]{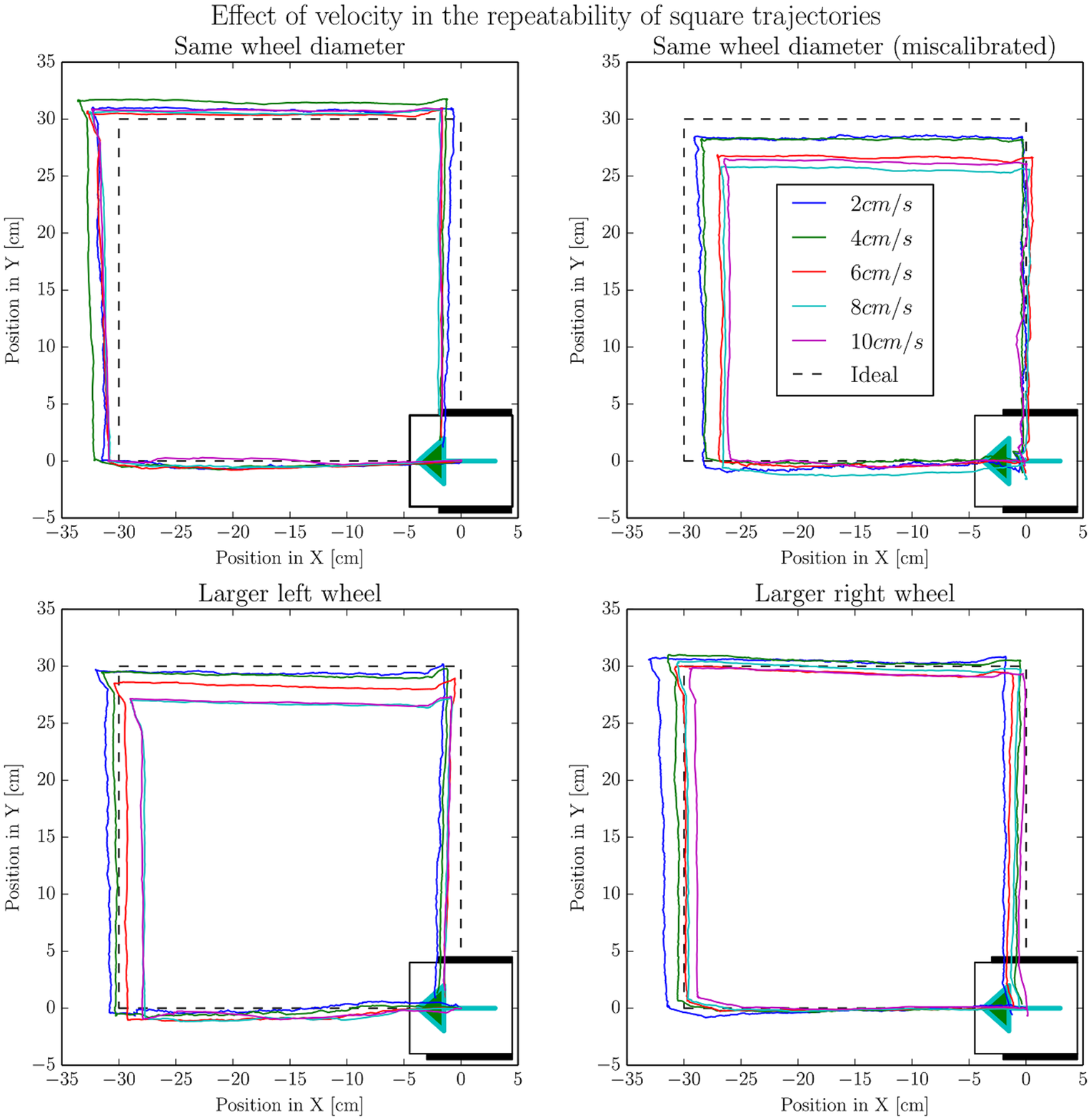}}
\caption[Effect of velocity in the repeatability of square trajectories]{\emph{Effect of velocity in the repeatability of square trajectories.}
The setup of these trials is similar to the one in Figure \ref{fig:square_distance_comparison}, but in this case the size of the squares were set to a constant $l$=$30cm$ and performed at different speeds.
The maximum drift is in the order of $6cm/m$ in the low velocity settings ($2$ to $6cm/s$) and around $9cm/m$ at higher velocities ($6$ to $10cm/s$).
}
\label{fig:square_velocity_comparison}
\end{figure}

\begin{figure}[hbtp]
\centerline{\includegraphics[width=1\linewidth]{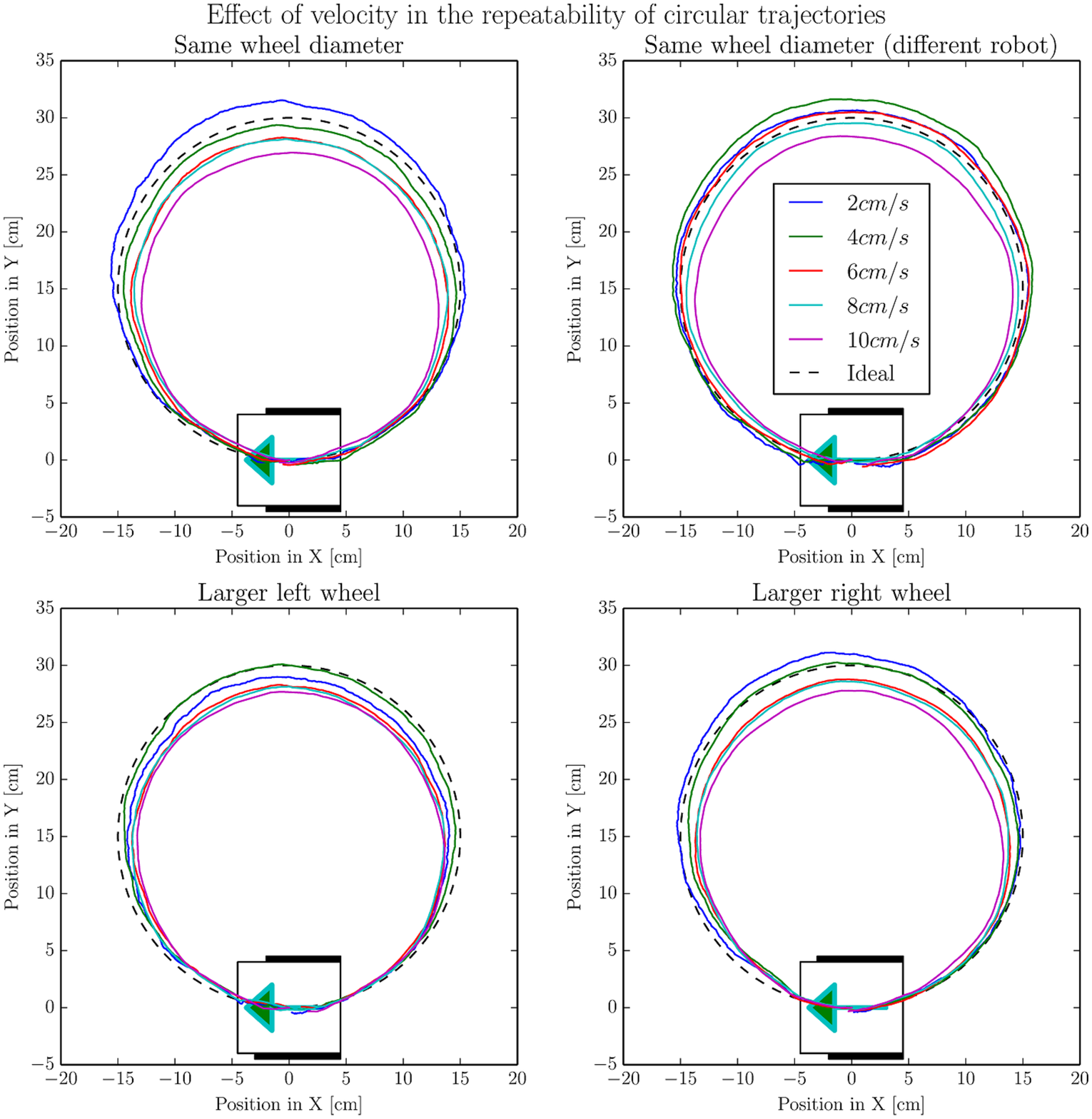}}
\caption[Effect of velocity in the repeatability of circular trajectories]{\emph{Effect of velocity in the repeatability of circular trajectories.}
The setup of these trials is similar to the one in Figure \ref{fig:circle_distance_comparison}, but in this case the circle diameters were set to a constant $D$=$30cm$ and performed at different speeds.
The maximum drift in diameter is around $7cm/m$ in the low velocity settings ($2$ to $6cm/s$) and around $8.5cm/m$ at higher velocities ($6$ to $10cm/s$).
}
\label{fig:circle_velocity_comparison}
\end{figure}

The implemented motion controller relies on the integration of velocities over time for its distance estimations.
In Section 2.4 it was observed that there is a nonlinear region in the velocity response curves (c.f. Figure \ref{fig:linear_velocity_calibration_curves}). Since these responses are linearly fitted, the use of velocities outside the linear range will result in inaccurate distance estimations. The effect of velocity is analysed in Figures \ref{fig:square_velocity_comparison} \& \ref{fig:circle_velocity_comparison}.

\newpage
\section{Performance in odor search experiments}

The GNBot robots were originally designed as a platform that could evaluate swarm search strategies in real-world environments\cite{GarciaSauraLM14,GarciaSaura14}. The target of those searches are odor sources based on volatile compounds (such as ethanol) that can be detected by the gas sensor on-board each robot.

Among the search strategies that have been so far evaluated with this platform are L\'{e}vy-based search (c.f. Fig. \ref{fig:levyDriftSingle}) and wall-bounce search.
Unfortunately, at the time of those trials a calibrated motion controller was not yet present. This fact caused great uncertainty in the evaluation process: the calculated motion commands for a particular search were not being faithfully reproduced by the swarm.

\begin{figure}[hbtp]
\centerline{\includegraphics[width=0.7\linewidth]{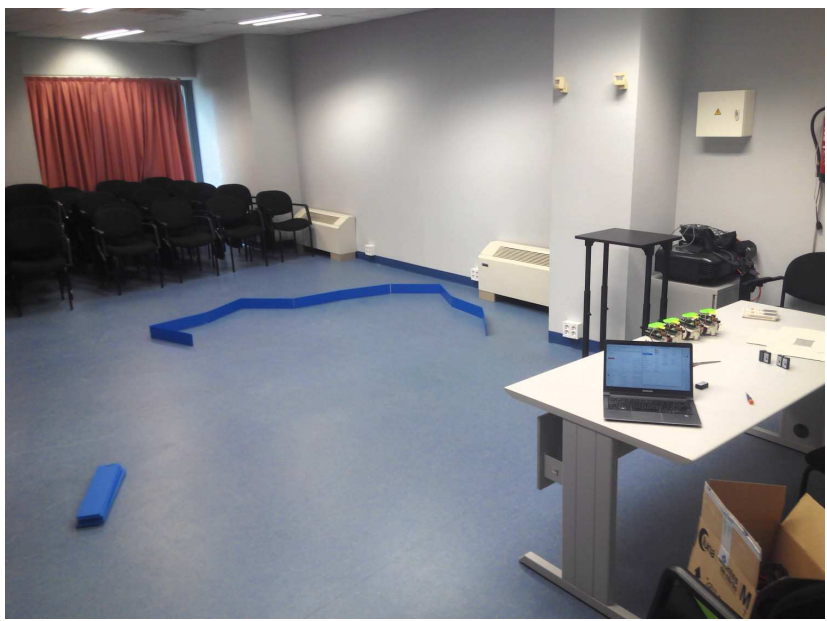}}
\caption[Room used for the odor search experiments]{\emph{Room used for the odor search experiments.} The odor search experiments required an effective area of approximately $16m^2$, an indoors environment free of disturbances and with uniform light conditions that facilitate video tracking.
Chairs were moved to the back of the room, and the search arena was installed in the center (blue walls). Air conditioning was turned off during the experiments on order to minimize air currents. The robot swarm can be seen on top of the table, as well as the computer used to record tracking data.
}
\label{fig:experiments_room}
\end{figure}

Some of those experiments have been re-run for this project -- this time using the self-calibrated motion controller in order to evaluate the improvement. Trials were performed in the seminar room shown in Figure \ref{fig:experiments_room}.

\begin{figure}[hbtp]
\centerline{\includegraphics[width=0.84\linewidth]{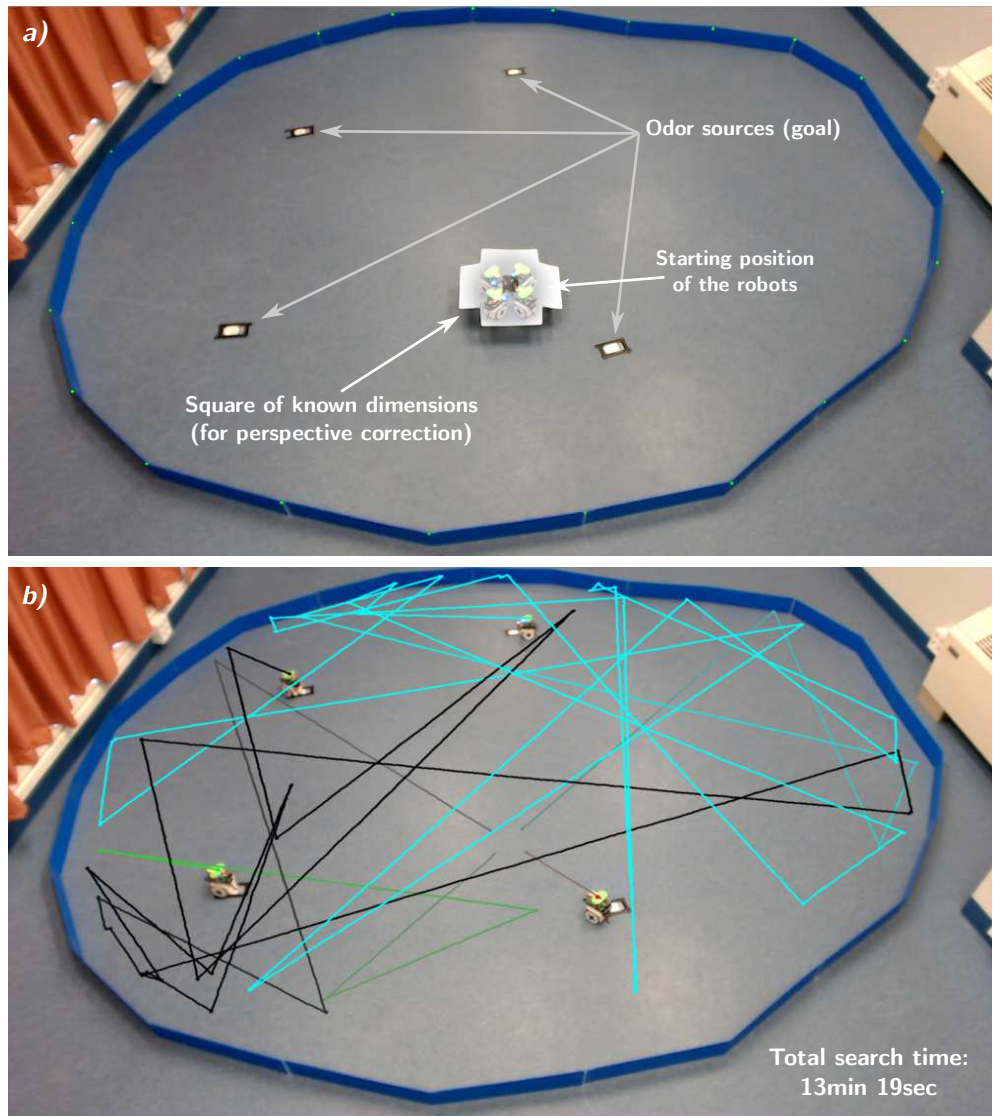}}
\caption[Setup of a wall-bounce search experiment (conducted with the new motion controller)]{\emph{Setup of a wall-bounce search experiment (conducted with the new motion controller).}
The upper panel shows the starting robot positions, as well as the location of the odor targets. These are based on cotton pads impregnated with ethanol. The lower panel shows an overlay of the search path described by each robot.
It can be appreciated that the trajectories are accurately performed as lines, which faithfully matches the high-level definition of the wall-bounce algorithm.
}
\label{fig:search_experiment_setup}
\end{figure}

Wall-bounce search is a very primitive form of search where a robot is commanded to maintain a constant velocity until either an obstacle or the target are encountered; for obstacles, the robot must rotate away and continue with a different heading.
The simplicity of the wall-bounce strategy makes it very sensible to the linearity of the robot's motions. For instance, a robot with enough yaw drift may rotate in circles  over the same area without ever finding any obstacle that modifies its path. Given its sensitivity to drift, wall-bounce search is a very good candidate for the comparison of both motion controllers.
The experimental setup is described in Figure \ref{fig:search_experiment_setup}.

The results of these search experiments are further discussed in the next section.

%%%%%%%%%%%%%%%%%%%%%%%%%%%%%%%%%%%%
\chapter{Conclusions} \label{sec:concl_chap}

This thesis has tackled the improvement of the motion controller on board the GNBot by incorporating gyroscope-based odometry that does not require wheel encoders.
The outcome is a high-level controller that can faithfully perform in-place rotations, arcs, and linear trajectories. This controller is calibrated, so its inputs are real-world units (centimeters and degrees). Additionally this calibration process is automatic and uses only two on-board sensing capabilities: the gyroscope and the distance sensor.

The general approach of the project has been to separately analyse the sources of motion uncertainty, and then try to find ways to minimise them (i.e. by observing a response curve and working only within its linear region). Many of the error sources could not be separated, so the calibration process had to progressively build a layered confidence base, which in the end allowed for complete self-calibration:

\vspace{-3.5mm}

\begin{itemize}
\item In first place, the output of the gyroscope (the MPU6050 inertial sensor) was analysed. It was found that the yaw measurement varied over time even when the robot was static, so this drift had to be compensated (Section \ref{sec:gyro_section}).
\item Then the response curves of the motors were analysed using the calibrated gyroscope. This was done by activating a single motor at a time and measuring the rotational velocities of the robot (c.f. Fig. \ref{fig:speed_calib_diagram}). The response curves were plotted for robots with different wheel diameters (c.f. Fig. \ref{fig:velocity_response_curve}). These were then linearised, which allowed control over real-world units of rotational speed (Section \ref{sec:rotVel_section}).
\item Using the calibrated motor functions, a PID yaw controller was implemented and tested in order to guarantee accurate yaw control (c.f. Fig. \ref{fig:P_controller_vs_autotuned_PID}). This allowed for fast and accurate yaw transitions. It also allowed for very accurate closed-loop linear trajectories (Section \ref{sec:PID_section}).
\item The IR distance sensor on-board the GNBot (Sharp GP2Y0A21YK0F) was then evaluated and linearised (c.f. Figs. \ref{fig:IR_sensor_response_curve_B} \& \ref{fig:IR_sensor_response_curve}).
Finally, the PID yaw controller was used in combination with the distance sensing capability in order to perform accurate velocity measurements. This way it was possible to measure the \emph{linear velocity response curves} of the robot (c.f. Fig. \ref{fig:linear_velocity_calibration_curves}). Basic fitting allowed for accurate velocity-based control with real-world input units (Section \ref{sec:linearVel_section}).
\end{itemize}

\vspace{-5mm}

The self-calibration process that has been implemented (c.f. Fig. \ref{fig:full_calibration_procedure}) uses only the gyroscope and distance sensors on board the GNBot to automatically measure all the necessary parameters, without requiring any user interaction. The process simply requires manual placement of the robot in front of a wall, and takes approximately 92 seconds.

\begin{figure}[hbtp]
\centerline{\includegraphics[width=1.1\linewidth]{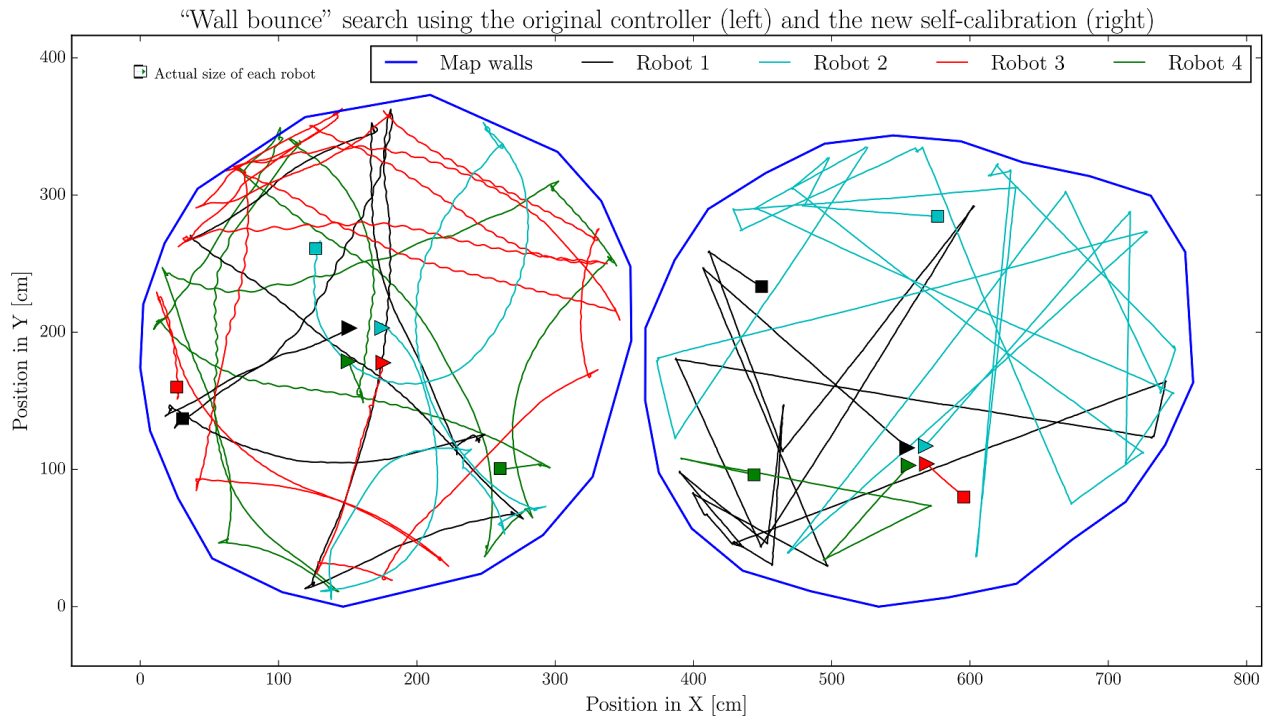}}
\caption[Comparison of a ``Wall bounce'' search using the original controller and the new self-calibration]{\emph{Comparison of a ``Wall bounce'' search using the original controller (left) and the new self-calibration (right).}
The figure shows the trajectories of four GNBot robots in two different odor search experiments. The starting positions are represented with triangle markers
and the targets are represented as squares.
It can be appreciated that the original controller (left panel) has large amounts of yaw drift. The new controller (right panel) faithfully matches the theoretical definition of wall-bounce search.
}
\label{fig:wallBounce_oldVSnew}
\end{figure}

Video tracking was used for the evaluation of the controller's accuracy (Chapter \ref{sec:eval_chap}). Tracking was implemented using the OpenCV library, first performing perspective correction, and then identifying the position of the color marker by applying a threshold (c.f. Figs \ref{fig:perspective_correction} \& \ref{fig:opencv_tracking_steps}). This tracking method was used to compare the new controller with the previously existing solution, in a wall-bounce odor search experiment (see Figure \ref{fig:wallBounce_oldVSnew}).

The tracking method also allowed to measure the effect of \emph{gyroscope drift} (by having the robot repeatedly perform straight lines, as seen in Section \ref{sec:sect31}), the \emph{linear positioning accuracy} (by performing squares and circles of different sizes, as seen in Section \ref{sec:sect32}), and the \emph{effect of velocity} in the controller (by performing squares and circles at different velocities, as seen in Section \ref{sec:sect33}).

The new motion controller has a linear drift in the order of $6cm/m$ in the low velocity settings ($2$ to $6cm/s$) and around $9cm/m$ at higher velocities ($6$ to $10cm/s$). The rotational drift is in the order of $2deg/min$ when the robot is moving.

In summary, this project has implemented the high-level functionality needed in order to achieve accurate motion control of the GNBot robots. The self-calibrating nature of the approach also facilitates its use in large robot swarms.
The outcome of the project has been published as open-source in a GitHub repository\footnote{\url{https://github.com/carlosgs/GNBot}}.

\section{Future work}

Using the new controller it will be possible to evaluate search strategies that depend on accurate position-based control (i.e. environment mapping, efficient area covering, flocking behaviours, etc).

In regards to self-calibration, the studied method provides static tuning parameters that remain unchanged until a re-calibration is manually issued.
Some authors have studied SLAM approaches that instead update the calibration estimates automatically in real time\cite{martinelli07,DeCecco02,roy99}.
These techniques could be explored with the incorporation of other forms of odometry such as wheel encoders.

The inertial sensor used in this project includes not only a gyroscope but also an accelerometer. It would be very interesting to incorporate these measurements into the motion controller. For instance, the accelerometer could be used in order to detect changes in ground surface and account for the differences in friction.

% Measure a different brand of servo motors/distance sensor (arduskybot)
% Incorporate the measurements of the accelerometer

\clearpage{\pagestyle{empty}\cleardoublepage}

%% bibliography
\renewcommand{\bibname}{References}
\newpage
\addcontentsline{toc}{chapter}{\bibname}
\bibliographystyle{unsrt}
\bibliography{references}

\end{document}